\documentclass[11pt]{article}
\newif\ifarxiv
\arxivtrue

  \usepackage{acl}

\usepackage{xurl}
\usepackage{times}
\usepackage{latexsym}
\usepackage[T1]{fontenc}
\usepackage[utf8]{inputenc}
\usepackage{inconsolata}
\usepackage{graphicx}
\usepackage{booktabs}
\usepackage{tabularx}
\usepackage{ragged2e}
\usepackage{array}
\usepackage{threeparttable}
\usepackage{enumitem}

\usepackage{amsmath}
\title{Framing by Wording, Framing by Selection: A Large-Scale Two-Dimensional Audit of French News Headlines, 2022--2025}

  \author{Amr Sobhy \\ Le French News Lab \\ \texttt{amr@frenchnewslab.org}}

\begin{document}
\maketitle

\setlength{\parskip}{5pt plus 1pt}

\begin{abstract}
  News headlines frame public issues both by \emph{what} they select and by \emph{how} they word it, yet computational framing work typically collapses these operations into a single score.
  We introduce a two-dimensional framework that separates \emph{salience framing}, measured through four wording devices (loaded vocabulary, blame attribution, threat framing, rhetorical question), from \emph{selection framing}, measured through outlet-level story-form and high-charge distributions.
  We build a 10,000-headline French supervision set using three LLM annotators with majority-vote resolution and human arbitration, validate the labels against two annotator-independent blind human studies, and apply the strongest classifier to 902,111 deduplicated headlines from 25 French outlets (2022--2025).
  Three main findings emerge.
  First, salience and selection divergence are positively correlated yet leave nearly half of outlet-level variance unexplained, populating interpretively distinct off-diagonal cells in a four-cell outlet typology.
  Second, default classification thresholds systematically inflate corpus-level salience estimates; a precision-floor recalibration protocol corrects this distortion.
  Third, group-mention analysis reveals sharply unequal salience contexts: headlines mentioning Jews, the Far-right, and Muslims carry the highest detected salience rates, which broad event-context composition does not fully explain (residuals are descriptive, not same-event causal estimates; per-group lexicon precision is reported alongside).
  To our knowledge, this is the largest framing-focused French \emph{headline} audit to date; we release the supervision set, lexicons, and analysis code.
  \end{abstract}
  
\section{Introduction}

Headlines are high-reach editorial artifacts: a majority of shared news URLs are never clicked \citep{gabielkov2016}, and experimental work shows that headlines can shape memory and inference even when the article is later read \citep{ecker2014}. Their compactness makes editorial choice unusually visible. A headline can frame by wording an event as invasion, betrayal, danger, scandal, or responsibility; it can also frame by repeatedly selecting particular event types for particular groups. When a French outlet writes \emph{``Les migrants envahissent''} (``Migrants are invading'') rather than \emph{``Des migrants arrivent''} (``Migrants are arriving''), the word choice is not accidental; it encodes a threat frame, an intergroup opposition, and a causal attribution in three words. When a far-right aggregator fills four in five headlines with crime stories about identifiable social groups, that selection pattern is not random; it systematically concentrates group-threat associations in editorial output. These are two distinct editorial operations.

Computational framing measures often blur them. Sentiment scores and bias classifiers capture how language is used but cannot distinguish neutral wording on a concentrated crime agenda from charged wording on a broader one; topic-distribution measures capture what is selected but not how it is packaged. Auditing harmful or polarizing media ecosystems, as the EU Digital Services Act's independent-audit regime now requires at platform scale, means knowing whether an outlet's effect is driven by selection, salience, or their interaction.
We study this problem in French news headlines, a setting with rich ideological variation but less computational framing infrastructure than English. We ask: \textbf{RQ1} how salience framing varies across French outlets; \textbf{RQ2} whether salience and selection are empirically dissociable, taking as the null that they are one construct measured twice, which would show as near-collinear divergences ($r \geq 0.90$) with unpopulated off-diagonal cells; and \textbf{RQ3} which social and political groups appear in the highest-salience headline contexts. Our contributions include the 10,000-headline supervision set, a precision-floor recalibration protocol, and a 902,111-headline audit showing substantial overlap between salience and selection without collapsing them into one axis.

\section{Framework and Related Work}
\label{sec:framework}

\citet{entman1993}'s definition says framing involves both \emph{selection} and \emph{salience}: selecting aspects of reality and making them more noticeable or meaningful. Entman treats these as coupled within a single framing act; we separate them not as a claim about the act but as a measurement decision at the outlet level, where the two need not coincide and, as \S\ref{sec:typology} shows, only partially do. We operationalize this as two measurable streams. \textbf{Salience framing} is the wording stream: four primary devices (loaded vocabulary \citep{stevenson1944}, blame attribution, threat framing, and rhetorical question) plus a supplementary indicator of ingroup/outgroup construction (us-vs-them). These devices are grounded in problem definition, causal interpretation, moral evaluation, and treatment implication, and in framing theory, critical discourse analysis, and argumentation theory \citep{entman1993,gamsonModigliani1989,vanDijk1991,reisiglWodak2001,walton1996,tankard2001}. The four primary devices are selected because they are theoretically grounded, observable in headline-length text, and stable enough for supervised measurement at corpus scale. The second criterion is the binding one and excludes devices the same traditions treat as central: personalisation and metaphor, for instance, are recoverable from article context but rarely decidable from a headline alone. \textbf{Selection framing} is the distributional stream: which story forms an outlet repeatedly chooses to cover, and how often those story forms are high-charge events such as crime, conflict, scandal, and crisis. This follows agenda-setting and gatekeeping traditions, where influence operates through coverage allocation and accessibility \citep{mccombsShaw1972,scheufeleTewksbury2007,shoemakerVos2009}. Our selection measure is conditional on coverage: it observes story-form distributions \emph{among headlines an outlet chose to publish}, closer to second-level attribute agenda-setting \citep{mccombs2005} than to first-level gatekeeping. Outlets with identical story-form distributions could differ substantially in which events from the world they select to cover at all: a dimension the published-headline corpus cannot observe.

Table~\ref{tab:schema} summarizes the measurement schema. The four primary devices are grounded in \citet{entman1993}'s four frame functions: blame attribution (causal interpretation), threat framing (problem definition and treatment implication), loaded vocabulary and rhetorical question (moral evaluation and problem definition), with many-to-many correspondence per headline; us-vs-them marks ingroup/outgroup opposition.
Prior work either assigns frame labels from a single typology without decomposing wording from selection \citep{card2015,liuFrames2019,mendelsohn2021}, captures a single bias dimension (informational and lexical bias \citep{fan2019,spinde2021} or persuasion techniques \citep{piskorski2023}), or separates issue filtering from slant rather than wording from selection \citep{budak2016}, and so cannot serve as quantitative baselines for a two-dimensional decomposition \citep{otmakhova2024}. English-language auditing resources including NELA-GT \citep{gruppi2022} and BERT-based political leaning classifiers \citep{baly2020} address related questions at scale but do not separate wording from selection mechanisms.

French-language media studies motivate the setting. \citet{bensonWood2015} documents that immigration news in France is dominated by governmental and political sources, with limited non-governmental voices \citep[cf.][for cross-national comparison]{benson2013}; \citet{dalibert2015} shows how minority-movement access to the French public sphere is regulated by ``francit\'{e},'' constraining visibility and legitimacy. Work on migrant representation records oscillation between victim, threat, and voiceless framings, echoing our threat and us-vs-them findings. This literature motivates the 12-group inventory analysed in \S\ref{sec:group_framing}, which includes migrants, Muslims, Jews, and the far right. Related francophone work detects opinion/information genre at scale in Qu\'{e}b\'{e}cois and Belgian media \citep{escouflaire2024} and links ownership structures to speaking-time-based political slant in French broadcasting \citep{cage2022}. More recent French-media framing and discourse studies remain either small and issue-specific, such as the 120-article immigration corpus of \citet{song2024}, or topic-bound at moderate scale, such as the 13,795-article AI press corpus of \citet{tsimpoukis2025}. Larger French news corpora, such as \citet{jehleLeGallo2025}'s 400,000-article EU-sentiment study, analyze tone or agenda dynamics rather than explicit framing measurement. Our unit is narrower, focusing on headlines only, but covers more outlets and explicitly separates selection from salience; to our knowledge, that makes the present corpus unusually large for framing-focused work in the French setting.

\begin{table}[t]
\centering
\scriptsize
\setlength{\tabcolsep}{3pt}
\renewcommand{\arraystretch}{1.12}
\begin{tabularx}{\columnwidth}{@{}l>{\RaggedRight\arraybackslash}X@{}}
\toprule
\textbf{Measure} & \textbf{Operational definition} \\
\midrule
Loaded vocabulary & Evaluative or emotionally charged wording whose neutral paraphrase would preserve the core event claim. \\
Blame attribution & Wording that assigns causal responsibility for a problem or negative outcome to an actor or group. \\
Threat framing & Wording that casts an actor, event, or situation as danger, invasion, crisis, or security risk. \\
Rhetorical question & Headline posed as a question. Operationalised by interrogative form, not by whether the question is judged rhetorical. \\
Us-vs-them & Explicit ingroup/outgroup contrast through naming, pronouns, identity markers, or oppositional structure. \\
Story form & 10-class event/action category: POLICY, PARLIAMENT, CONFLICT, CRIME, SCANDAL, ELITE, SOCIAL, LAW, ELECTIONS, OTHER. \\
High-charge & Binary flag for story forms associated with conflict, crime, scandal, crisis, or strong negativity. \\
\bottomrule
\end{tabularx}
\caption{Measurement schema. Salience devices are wording-level and non-mutually exclusive; story form and high-charge are selection-level.}
\label{tab:schema}
\end{table}

The boundary matters most for attribute agenda-setting \citep{mccombs2005}: repeated CRIME coverage operates through salience transfer and accessibility, making considerations available in memory \citep{mccombsShaw1972,scheufele1999}, while charged wording operates through applicability, shaping which interpretive schemas audiences invoke \citep{scheufele1999,scheufeleTewksbury2007}. A single ``framing intensity'' score cannot preserve that distinction.

\setlength{\parskip}{0pt}%
\section{Data and Method}
\label{sec:data_method}

\paragraph{Corpora.}
We use two datasets with different roles. The supervised development set contains 10,000 France-based news headlines from 25 outlets, designed as a structurally comprehensive cross-section of the French media ecosystem across the political spectrum \citep{newman2023reuters,acpm2026}. It includes three of France's leading broadcast and continuous-news brands (\textit{Franceinfo}, \textit{BFMTV}, \textit{TF1 INFO}) and seven major national dailies (\textit{Le Monde}, \textit{Le Figaro}, \textit{Le Parisien}, \textit{Les Echos}, \textit{Lib\'{e}ration}, \textit{La Croix}, \textit{L'Humanit\'{e}}). Selection criteria were reach or circulation rank; format diversity; ideological breadth, explicitly anchoring the range with \textit{La Croix} (Catholic daily), \textit{L'Humanit\'{e}} (communist historical title), \textit{Fdesouche} (far-right aggregator), \textit{Valeurs actuelles} (conservative-nationalist weekly), and \textit{Causeur} (conservative intellectual magazine); and restriction to outlets covering three shared sections, with headlines jointly balanced across outlets and sections.

Each headline is labeled by three schema-constrained LLM annotators from different model families, following evidence that LLMs can match or outperform crowdworkers on structured annotation tasks \citep{gilardi2023,tornberg2023,ziems2024}---a task-specific comparison that remains contested, and one our own validation qualifies (Table~\ref{tab:human_agreement}). The annotators, \texttt{openai/gpt-oss-120b} (released August 2025), \texttt{google/gemma-4-31B} (released April 2026), and \texttt{meta-llama/Llama-3.3-70B-Instruct}, produced full-field labels under a shared schema-constrained instruction set; Appendix Table~\ref{app:prompt} summarizes the released schema; the full annotation prompt is included in the release. Three candidate second annotators were run against the primary annotator at full scale and two were retained; pairwise agreement statistics for all three are released. Final binary labels are resolved by 2/3 majority vote. The 642 three-way story-form conflicts (6.4\%) are assigned to a single human arbitrator; post-hoc blind inter-annotator reliability on these cases is reported in Appendix Table~\ref{app:arbitration_iaa}. Binary conflicts are not arbitrated: the panel is three models with overlapping pretraining rather than three independent coders, so majority vote can propagate shared error (Appendix Table~\ref{app:homogeneity}), and the blind human studies rather than arbitration are the control for the binary heads; the weakest head, us-vs-them, is reported only as a supplementary indicator. The final split is 6,999 train, 1,500 validation, and 1,501 test headlines.

The corpus-level analysis set approximates real editorial output. We drew a proportionally stratified sample targeting one million headlines from the 25-outlet panel across outlet, section, and year for 2022--2025, then removed exact duplicate URLs and within-outlet duplicate headlines, yielding 902,111 unique headlines. Eligibility was restricted to the Politics, Economy, and Society sections used throughout corpus construction. Per-outlet counts range from 646 (Blast) to 108,307 (Le Figaro); full counts appear in Appendix Table~\ref{app:outlets}. Identical headlines across outlets are retained (syndication is part of the observable editorial field); classification is restricted to headline text, the standalone framing unit. The unit also strips context that disambiguates attribution: quotation, irony, and blame voiced by a cited source rather than by the outlet are often unrecoverable, which bounds the blame-attribution and threat-framing devices in particular.

\paragraph{Labels.}
Salience is multi-label: loaded vocabulary, blame attribution, threat framing, rhetorical question, and us-vs-them construction. Selection is modeled through a binary high-charge label \citep{galtungRuge1965,harcupOneill2017} and a 10-class story-form taxonomy (Table~\ref{tab:schema}), developed iteratively during pilot annotation on a 500-headline exploratory sample and guided by French press section conventions and prior news-values classifications \citep{harcupOneill2017}; unlike topic-oriented systems such as IPTC Media Topics, story form classifies the journalistic action a headline reports rather than just the subject domain, so one immigration headline may be POLICY and another CRIME (Appendix Table~\ref{app:device_examples}). Target-group analysis is handled separately through a hand-curated lexicon of 219 French surface-form terms spanning 12 groups (Appendix Table~\ref{tab:label_distribution} reports the full \texttt{v6} label distribution).

\paragraph{Models.}
We compare three supervised measurement stacks on the final \texttt{v6} split: a TF-IDF + Logistic Regression baseline, \texttt{camembert-base}, and \texttt{xlm-roberta-large} (Table~\ref{tab:model_comparison}). For transformer models, a multi-task encoder predicts the five salience devices and charge band through six independent binary heads on a shared headline representation, while a separate encoder predicts story form through a single 10-class head. We use \texttt{xlm-roberta-large} for corpus inference because it delivers the strongest overall held-out performance: mean salience+charge F1 is .742 (XLM-R), .726 (CamemBERT), .476 (TF-IDF); XLM-R also leads on story-form macro-F1 (.731). CamemBERT scores higher on rhetorical questions alone (.900 vs.\ .841), likely reflecting French-specific interrogative morphosyntax; for corpus inference we therefore use XLM-R for all heads except rhetorical question, where CamemBERT is substituted. The ensemble was fixed on validation-set metrics before corpus inference; no post-hoc adjustments were made. Exhaustive encoder benchmarking is not the contribution; the substantive cross-outlet findings are bounded against encoder choice by the TF-IDF replication in \S\ref{sec:robustness} ($r=0.667$ vs.\ $0.736$).

Both models use a 48-token maximum sequence length ($<$2\% of headlines exceed this; details in Appendix Table~\ref{app:hyperparams}). Three-seed evaluation confirms robust model selection (Appendix Table~\ref{app:hyperparams}); results in Table~\ref{tab:model_comparison} are from seed~42. Thresholds are selected on validation only via a precision-floor-constrained recalibration policy: for each binary head, we search a 0.10--0.99 grid and retain the threshold maximizing validation F1 subject to pre-specified per-head precision floors, assigned from annotator agreement tiers before validation metrics are observed: rhetorical question ($\kappa{=}.770$, highest) receives a .85 floor; the mid-agreement heads (loaded vocabulary, blame attribution, charge band) receive .70; and the lowest-agreement heads (threat framing, us-vs-them) receive .60 (per-head floors and thresholds in Table~\ref{tab:thresholds}).

\begin{table}[t]
\centering
\scriptsize
\setlength{\tabcolsep}{2.2pt}
\renewcommand{\arraystretch}{1.1}
\begin{tabular*}{\columnwidth}{@{\extracolsep{\fill}}lrrrrr@{}}
\toprule
\textbf{Head} & \textbf{MinP} & \textbf{Thr.} & \textbf{P} & \textbf{R} & \textbf{F1} \\
\midrule
loaded & .70 & .67 & .733 & .698 & .715 \\
blame & .70 & .83 & .719 & .650 & .683 \\
threat & .60 & .84 & .723 & .712 & .718 \\
rhet. q. & .85 & .89 & .957 & .809 & .877 \\
\midrule
us-vs-them$^{\ddagger}$ & .60 & .89 & .600 & .625 & .612 \\
charge & .70 & .64 & .779 & .785 & .782 \\
any sal. & -- & -- & .838 & .780 & .808 \\
\bottomrule
\end{tabular*}
\caption{Final \texttt{v6} thresholds (validation set only) for the mixed-encoder ensemble; see Table~\ref{tab:model_comparison} and Models. $^{\ddagger}$Us-vs-them is a supplementary indicator (below the $0.60$ precision floor; Appendix Table~\ref{app:uvt_supplement}), not a primary prevalence estimate.}
\label{tab:thresholds}
\end{table}

\begin{table*}[t]
\centering
\scriptsize
\setlength{\tabcolsep}{4pt}
\renewcommand{\arraystretch}{1.12}
\begin{tabular*}{\textwidth}{@{\extracolsep{\fill}}lccccccc@{}}
\toprule
\textbf{Model} & \textbf{Loaded} & \textbf{Blame} & \textbf{Threat} & \textbf{Rhet.Q} & \textbf{UvT}$^{\ddagger}$ & \textbf{Charge} & \textbf{Story (macro-F1)} \\
\midrule
TF-IDF + Logistic Regression & .518 & .468 & .429 & .491 & .381 & .572 & .543 \\
CamemBERT-base & .770 & .657 & .642 & .900 & .617 & .769 & .723 \\
XLM-RoBERTa-large$^\dagger$ & .758 & .672 & .749 & .841 & .633 & .801 & .731 \\
\bottomrule
\end{tabular*}
\caption{Held-out test-set F1 by model (final \texttt{v6} split, seed~42, production thresholds); per-model means and three-seed robustness are in Models and Appendix Table~\ref{app:hyperparams}. $^\dagger$Production model; CamemBERT substituted for rhetorical question only (see Models). $^{\ddagger}$Us-vs-them is a supplementary indicator (below the $0.60$ precision floor; Appendix Table~\ref{app:uvt_supplement}), not a primary prevalence estimate.}
\label{tab:model_comparison}
\end{table*}

Agreement patterns (Appendix Table~\ref{tab:agreement}) are consistent with the \texttt{v6} merge design: binary fields show high unanimous agreement while story form is harder, matching the need for arbitration; us-vs-them is weakest ($\kappa=.408$).

\paragraph{Independent blind validation.}
Two annotators with no affiliation to this work independently re-annotated a stratified 499-headline held-out sample (primary validation) using a bilingual annotation guide derived from theoretical construct definitions, with no access to LLM labels or expected label distributions (Table~\ref{tab:human_agreement}). The sample is temporally representative of the held-out test split (39.5\% post-October 7, 2023, vs.\ 41.3\% of date-resolved test headlines), directly covering the period most exposed to possible LLM pretraining overlap; within this subset, any-salience human--LLM agreement declines only modestly ($\kappa=.424$ vs.\ $.463$ full-sample). Majority-vote human--LLM $\kappa$ ranges $.608$--$.869$ across the four primary devices; recall is consistently high ($\geq .65$) while precision is lower ($.37$--$.82$), confirming that the merged labels are systematically liberal (see Limitations). Human--human $\kappa=.680$ on the any-salience composite confirms that annotators agree on overall salience judgements even where device attributions diverge. A corroborating independent study on a separate $N{=}350$ sample yields consistent results (Appendix Table~\ref{tab:indep_validation}). Appendix Table~\ref{tab:error_analysis} provides a qualitative error analysis illustrating principal failure modes.

\begin{table}[t]
\centering
\scriptsize
\setlength{\tabcolsep}{2.2pt}
\renewcommand{\arraystretch}{1.08}
\begin{tabular*}{\columnwidth}{@{\extracolsep{\fill}}lrrrrr@{}}
\toprule
\textbf{Device} & \textbf{H--H $\kappa$} & \textbf{Maj--LLM $\kappa$} & \textbf{F1} & \textbf{Prec.} & \textbf{Rec.} \\
\midrule
Loaded vocabulary   & .542 & .608 & .768 & .644 & .951 \\
Blame attribution   & .601 & .718 & .675 & .579 & .809 \\
Threat framing      & .613 & .668 & .564 & .420 & .857 \\
Rhetorical question & .894 & .869 & .862 & .820 & .909 \\
\midrule
Us-vs-them\textsuperscript{†} & .510 & .596 & .471 & .369 & .649 \\
Any salience\textsuperscript{‡} & .680 & .463 & .780 & .646 & .984 \\
\bottomrule
\end{tabular*}
\caption{Independent blind re-annotation of the 499-headline validation sample ($N=499$) by two unaffiliated annotators. Maj--LLM = majority-vote human labels vs.\ LLM consensus; conflict rows excluded, effective $N$ per head 383--488. $\kappa$ interpretation bands follow \citet{landisKoch1977}. \textsuperscript{†}Us-vs-them is a supplementary indicator (below the $0.60$ precision floor; see Appendix Table~\ref{app:uvt_supplement}), not a primary prevalence estimate. \textsuperscript{‡}Derived composite. See Appendix Table~\ref{tab:indep_validation} for the corroborating $N=350$ study.}
\label{tab:human_agreement}
\end{table}

Group analysis uses a lexicon of explicit anchor terms mapping French surface forms to 12 canonical groups. The lexicon contains 219 terms and detects 134,046 headlines (14.9\% of the corpus), covering explicit mentions only; indirect or paraphrastic references are missed by design. Groups were selected by theoretical relevance (targeting identities whose framing asymmetries are documented in French and European media research \citep{bensonWood2015,dalibert2015,vanDijk1991}) and by a minimum corpus-frequency criterion of at least 1,000 detectable headline mentions in a pilot pass. A blinded two-human audit of a 200-headline sample confirms high agreement: $\kappa=.94$--$.98$ on the representative tier ($n{=}140$) and $\kappa=.83$--$.85$ on a harder stress-test tier ($n{=}60$). The lexicon shows precision $\geq .83$ for 9 of 12 groups; LFI (.69), Jews (.70), and Unions (.82, marginal) are lower-precision and should be treated as indicative (Appendix Table~\ref{tab:group_validation}).

\begin{figure*}[!t]
\centering
\includegraphics[width=\textwidth]{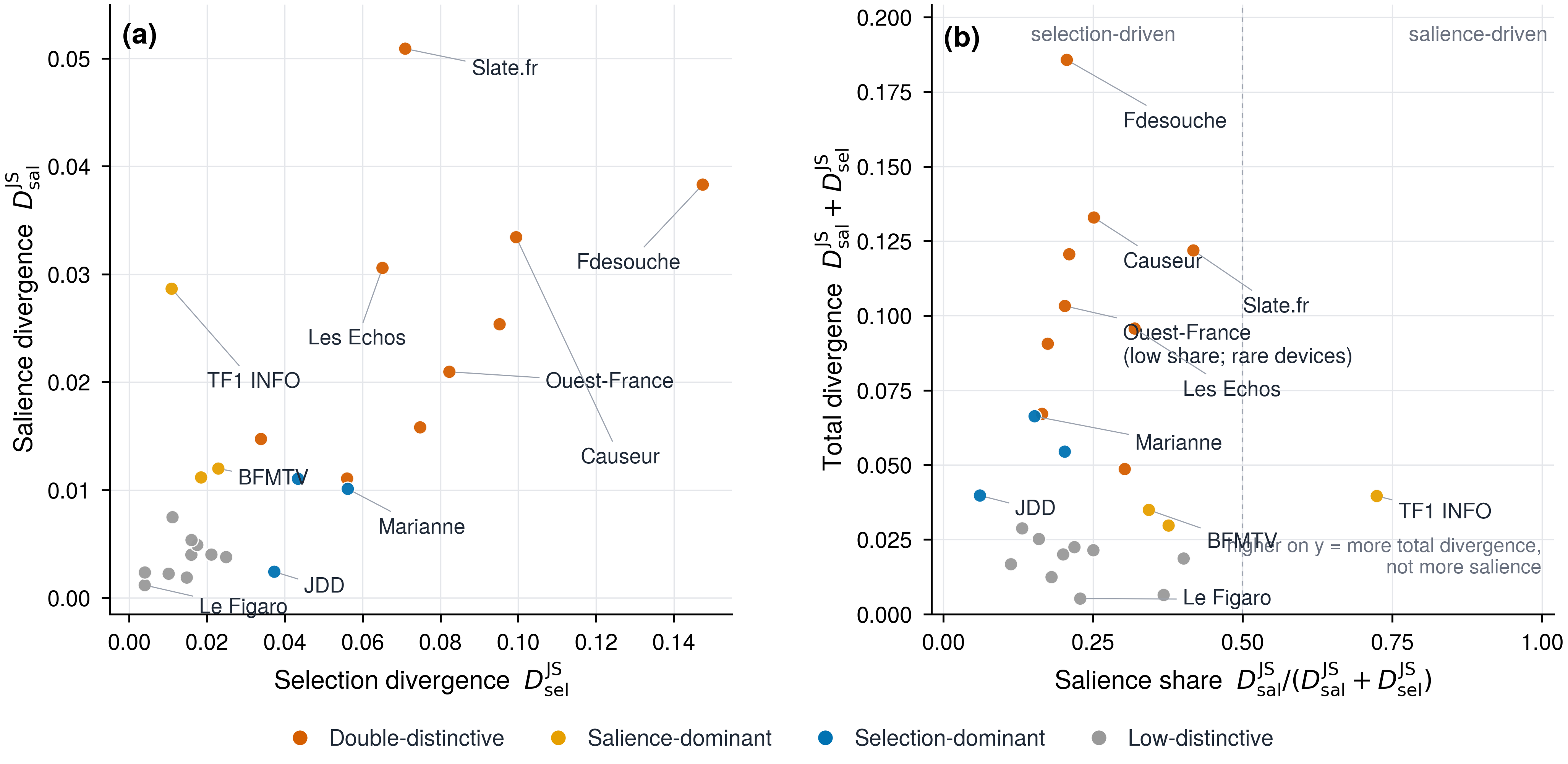}
\caption{\textbf{Selection vs.\ salience divergence across French
news outlets.} Note that \emph{Ouest-France} sits high in (b) because it is
far from the baseline overall while its headlines deploy framing devices
unusually \emph{rarely}, not because they deploy them aggressively.
(a)~Each outlet's distance from the corpus baseline decomposed into selection
divergence ($D_\mathrm{sel}^\mathrm{JS}$, what stories are emphasised) and
salience divergence ($D_\mathrm{sal}^\mathrm{JS}$, which framing devices are
deployed in the headline).  Both quantities are Jensen--Shannon divergences
against the cross-outlet mixture distribution.  Marker colour encodes the
four-class typology defined in \S\ref{sec:typology}.
(b)~Same outlets re-expressed as a salience share
($D_\mathrm{sal}^\mathrm{JS} / (D_\mathrm{sal}^\mathrm{JS} +
D_\mathrm{sel}^\mathrm{JS})$) against summed divergence.  The dashed line at
$0.5$ separates outlets whose share falls on the selection side from those on the
salience side.  Vertical position in (b) is summed distance from the baseline,
not salience intensity; outlets close to the bottom edge are close to the corpus
baseline on both dimensions.  Both divergences share the same bounded scale, so
the share is well defined, but their expected magnitudes differ with support size
(5 vs.\ 10 bins); both coordinates in (b) are therefore read ordinally --- as
which axis dominates within an outlet, and as rank distance from the baseline ---
not as an equivalence of magnitudes.}
 \label{fig:outlet_strategy}
\end{figure*}

\paragraph{Analysis plan.}
We summarize salience and selection distinctiveness with Jensen--Shannon divergence from the corpus baseline, following the use of Jensen--Shannon measures for media-agenda comparison \citep{pinto2019}, and test dissociability with Pearson and Spearman correlations against the conventional $r<0.90$ criterion (RQ2). We build the four-cell typology from median splits on salience and selection divergence as a cartographic convenience rather than a claim of categorical differences; continuous outlet positions are visible in Figure~\ref{fig:outlet_strategy}. Robustness checks compare focal outlets against all other outlets within the same story form using $\chi^2$ and $\phi$ as a coarse descriptive stress test; story form is partly editorially mediated and therefore is not an exogenous same-event control. Salience distinctiveness measures divergence in the distribution of rhetorical devices relative to the corpus baseline rather than raw device frequency, so an outlet may exhibit frequent charged language while remaining low in Sal.JS if its device mixture mirrors the broader media distribution. Because Sal.JS is computed over a 5-bin device distribution and Sel.JS over a 10-bin story-form distribution, the two divergences are not directly comparable in magnitude; the typology uses within-axis rankings rather than cross-axis absolute values.
\section{Results}

Across the full corpus, 34.6\% of headlines contain at least one salience device (loaded vocabulary most frequent, us-vs-them rarest; Appendix Table~\ref{tab:label_distribution}). High-charge story forms account for 31.3\%. These rates are lower than default-threshold estimates would produce, reflecting the precision-floor recalibration; the 34.6\% figure is best interpreted relative to the French corpus baseline rather than as a universal prevalence benchmark.

\subsection{Outlet Profiles and Dissociability}
\label{sec:typology}

Figure~\ref{fig:outlet_strategy} maps all 25 outlets in the two-dimensional divergence space, Table~\ref{tab:typology} summarizes the four-cell membership, and Appendix Table~\ref{app:outlets} reports the full outlet-level values. Salience divergence from the corpus baseline varies sharply: Slate.fr is the most salience-distinctive outlet, Fdesouche carries the clearest threat/blame profile, Causeur is distinctive through loaded and interrogative wording, and Ouest-France is double-distinctive through unusually low device and high-charge rates (7.2\% any-salience, 4.6\% high-charge, both lowest in the panel). The Double-distinctive cell spans 7.2\%--72.6\% any-salience: cell membership reflects position relative to thresholds, not editorial similarity within the cell. Selection divergence produces a different ranking, with Fdesouche, Causeur, and Blast among the clearest agenda outliers (Causeur's ELITE-heavy selection profile should be read with caution because ELITE-category conflict cases show lower arbitration agreement; Appendix Table~\ref{app:arbitration_iaa}) and high-charge coverage concentrated especially in Fdesouche, Blast, and Valeurs actuelles.

Device composition reveals three outlet signatures: \textit{threat-blame}, where threat and blame co-elevate (Fdesouche most clearly, with Valeurs actuelles at lower intensity); \textit{interrogative-evaluative}, where rhetorical questions and loaded vocabulary dominate without the broad threat/blame profile (Slate.fr, Causeur, L'Express; rhetorical-question rates index headline form, see Limitations); and \textit{institutional blame}, where accountability language rises without a strong threat register (Mediapart, L'Humanit\'{e}, Le Parisien; blame rates are similarly directional). (Loaded vocabulary: est.\ precision $0.643$ post-prior-shift, $-5.7$\,pp below the $0.70$ floor; Appendix Table~\ref{app:threshold_drift}.) Similar any-salience rates can therefore reflect distinct interpretive mechanisms.

Rate estimates widen materially only for the smallest outlets; Blast (646 headlines) should be read as an illustrative case profile (Appendix Table~\ref{tab:ci_audit}).

\begin{table}[t]
\centering
\scriptsize
\setlength{\tabcolsep}{3pt}
\renewcommand{\arraystretch}{1.12}
\begin{tabular*}{\columnwidth}{@{\extracolsep{\fill}}lrrrr@{}}
\toprule
\textbf{Cell} & \textbf{$n$} & \textbf{Sal.JS} & \textbf{Sel.JS} & \textbf{Sal.\%} \\
\midrule
Double-distinctive & 9 & 0.027 & 0.081 & .484 \\
Salience-dominant & 3 & 0.017 & 0.018 & .358 \\
Selection-dominant & 3 & 0.008 & 0.046 & .510 \\
Low-distinctiveness & 10 & 0.004 & 0.014 & .376 \\
\bottomrule
\end{tabular*}
\caption{Four-cell typology. Sal.JS/Sel.JS = salience/selection Jensen--Shannon divergence from the corpus baseline (§\ref{sec:data_method}); cells assigned by median splits; Sal.\% = mean any-salience prevalence (selection-dominant outlets show the highest Sal.\% yet the lowest Sal.JS; see Results). Outlet assignments in Fig.~\ref{fig:outlet_strategy} and Appendix Table~\ref{app:outlets}.}
\label{tab:typology}
\end{table}

Dissociability rests directly on the populated off-diagonal cells of the bivariate distribution (Figure~\ref{fig:outlet_strategy}; four-cell summary in Table~\ref{tab:typology}): the cells are perturbation-stable (only seven near-boundary outlets shift under $\pm$20\% median changes, Appendix Table~\ref{tab:typology_perturbation}), and concrete cases anchor the corners (JDD high on selection yet low on salience, TF1 INFO the reverse), holding independently of any correlation threshold. The correlation only quantifies the overlap: salience distinctiveness explains 54.2\% of outlet-level selection variance ($r{=}0.736$\footnote{Split-half reliability: $r_{xx}{=}0.979$, $r_{yy}{=}0.988$; Spearman-corrected latent $r{=}0.748$ (see Limitations for the leave-one-out baseline correction; $r{=}0.721$).}, LOO-corrected $r{=}0.721$; 95\% bootstrap CI [.503, .936], $N{=}25$; reliably non-zero, permutation $p{=}.0001$), leaving 45.8\% unexplained. Both estimates fall within the conventional $r<0.90$ criterion; with $N{=}25$ the interval is wide and its upper bound ($.936$) exceeds it, so the claim does not rest on the threshold. This residual maps onto interpretable off-diagonal profiles: selection-dominant outlets exhibit the \textit{highest} raw salience prevalence (any-salience~$=0.510$) despite the \textit{lowest} rhetorical divergence (Sal.JS~$=0.008$), while double-distinctive outlets combine high salience (any-salience~$=0.484$) with substantially greater divergence (Sal.JS~$=0.027$). Frequent charged language therefore does not by itself imply rhetorical distinctiveness.

To test whether this association is ecological aggregation, we estimate a longitudinal outlet$\times$month panel (1,184 observations; Appendix Table~\ref{app:panel_sensitivity}). Under two-way outlet+month fixed effects with cluster-robust inference (outlet clusters, $G{=}25$), the within-outlet coupling remains positive ($\beta=0.758$, 95\% CI [$-$.11, 1.63]; $p=.084$); we treat this as confirmatory rather than a stand-alone test. Excluding Fdesouche lowers the cross-sectional correlation to $r=0.688$ ($\rho=0.711$), and the unexplained variance ($1{-}r^2=0.53$) remains sufficient to populate distinct off-diagonal cells. Because three high-volume low-divergence outlets (Le Figaro, Le Parisien, Franceinfo) contribute ${\approx}$32\% of the corpus baseline, their low divergence is partly self-referential; a leave-one-out correction yields $r{=}0.721$, $\rho{=}0.727$, preserving the dissociability criterion (full LOO analysis in Limitations). A tone-only audit would therefore miss selection-dominant outlets, while a topic-only audit would miss salience-dominant ones.

\subsection{Group-Mention Salience Contexts}
\label{sec:group_framing}

\citet{vanDijk1991} establishes that group framing operates through both event-context selection and adversarial wording devices; our two-dimensional framework keeps those pathways analytically separate. Explicit group mentions occur in highly asymmetric model-estimated salience contexts (Table~\ref{tab:groups_main}). Group salience rates reflect headline rhetorical intensity, not editorial sentiment or targeting; interpretive caveats in \S{}Ethics apply throughout. Headlines mentioning Jews have the highest observed any-salience rate, driven predominantly by antisemitism reporting and Israel/Gaza security coverage (lexicon precision for this group is .70, among the lowest of the twelve, though recall is .97, so the detector is liberal rather than blind): 78.7\% carry at least one detected salience device and 79.8\% are high-charge. A year$\times$story-form reweighting estimates that 49.6\% salience would be expected from their coarse event-context mix alone, leaving a descriptive +29.1\,pp residual; this residual is evidence that broad story-form composition does not fully explain the pattern, not evidence of same-event causal editorial targeting. Far-right mentions show a similar raw/residual profile (77.7\% raw; +27.7\,pp residual), reflecting electoral, parliamentary, and conflict coverage of a political movement rather than targeting of its members. Muslims (66.4\%; +19.8\,pp residual) and Migrants (58.4\%; +17.7\,pp) concentrate in policy and security event contexts; Police (57.6\%; +11.1\,pp) coverage is dominated by blame attribution in accountability-focused reporting; and Unions (53.8\%; +3.6\,pp) reflect the conflict-heavy 2023--2024 pension-reform mobilization cycle. Seniors remain below baseline, and Workers fall below the salience rate expected from their year$\times$story-form mix. The 2024--2025 salience rise for Jews (77.2\% to 84.1\%; Appendix Table~\ref{tab:yearly_groups}) is consistent with post-October-2023 crisis coverage; high salience reflects event-selection and wording contexts, not a uniform targeting pattern (caveats in \S{}Ethics).

\begin{table}[t]
\centering
\scriptsize
\setlength{\tabcolsep}{2.5pt}
\renewcommand{\arraystretch}{1.08}
\begin{tabular*}{\columnwidth}{@{\extracolsep{\fill}}lrrrrr@{}}
\toprule
\textbf{Group} & \textbf{N} & \textbf{Sal\%} & \textbf{Exp\%} & \textbf{Res.} & \textbf{Top form} \\
\midrule
Jews & 4,646 & 78.7 & 49.6 & +29.1 & CRIME \\
Far-right & 3,981 & 77.7 & 50.0 & +27.7 & CONFLICT \\
Muslims & 4,720 & 66.4 & 46.6 & +19.8 & CRIME \\
Migrants & 9,606 & 58.4 & 40.7 & +17.7 & CRIME \\
Police & 14,313 & 57.6 & 46.5 & +11.1 & CRIME \\
Unions & 6,877 & 53.8 & 50.3 & +3.6 & CONFLICT \\
Workers & 15,456 & 39.8 & 52.8 & $-$13.0 & CONFLICT \\
Seniors & 21,987 & 32.1 & 34.6 & $-$2.5 & POLICY \\
\bottomrule
\end{tabular*}
\caption{Group-mention salience contexts (top 8 of 12 groups). Sal\% = observed any-salience; Exp\% = expected rate under year$\times$story-form reweighting (focal group excluded from cell baselines to avoid circular self-inclusion); Res. = observed $-$ expected (pp). Residuals are descriptive, not same-event causal estimates. Full 12-group table in Appendix Table~\ref{tab:groups}.}
\label{tab:groups_main}
\end{table}

All differences from baseline are significant under BH-FDR-corrected two-proportion $z$-tests (12 hypotheses; $p < 0.001$; intervals do not propagate classifier uncertainty; see Limitations for causal-identification disclosures).

Device profiles differ by group (Appendix Table~\ref{tab:group_mechanisms}): Jews show the strongest multi-device signal (blame, loaded vocabulary, threat); Police are blame-dominated; and the Far-right shows broad elevation across all three adversarial devices. Selection adds a second layer: Police and Migrants concentrate in CRIME; Workers and Unions concentrate in CONFLICT (56.9\% and 56.8\%, reflecting the 2023--2024 pension-reform cycle); and RN concentrates in ELECTIONS and PARLIAMENT.

Outlet-by-group patterns (Appendix Table~\ref{tab:group_mechanisms}) add an important caution: Fdesouche's elevation spans multiple groups, indicating a broad high-intensity editorial register rather than targeted group framing, while Mediapart and L'Humanit\'{e} peak on Police and the Far-right through an accountability lens. Removing Fdesouche confirms most group effects are robust (median exclusion effect $-$0.5\,pp; Appendix Table~\ref{app:fde_exclusion}). Results for the supplementary us-vs-them indicator are reported in Appendix Table~\ref{app:uvt_supplement}.

\subsection{Robustness}
\label{sec:robustness}

A dual-ablation study confirms that predictions are not driven by entity memorization: named-entity masking (spaCy) yields negligible confidence drops ($|\Delta p| \le 0.085$) and low label-flip rates (5.3\%--11.4\%), while masking the single most salient semantic token (Integrated Gradients) produces substantially larger effects (e.g., $|\Delta p| = 0.266$, 31.6\% flip rate for rhetorical questions). Within-story-form comparisons show that salience differences persist inside broad story-form bins (Appendix Table~\ref{tab:within_form}): Fdesouche and Valeurs actuelles remain elevated within CRIME, and Slate.fr retains its interrogative-evaluative profile within SOCIAL; these are descriptive stress tests rather than causal controls because story-form assignment is itself partly shaped by editorial judgment. A within-outlet headline bootstrap confirms near-perfect rank stability ($\rho{=}.996/.998$ for salience/selection JS), and selection-divergence rankings are robust to taxonomy granularity (Appendix Table~\ref{app:rq_excluded}). The top group pattern is temporally persistent: Jews, the Far-right, and Muslims remain the three highest-salience groups in every year from 2022 to 2025, with annual lifts in the $\times$1.78--$\times$2.35 range (Appendix Table~\ref{tab:yearly_groups}).

An encoder-independence check recomputes Sal.JS/Sel.JS with a TF-IDF story-form model (macro-F1=0.543): $r$ drops from 0.736 to 0.667 (still $<0.90$); $\rho$ is stable (0.741 vs.\ 0.747), bounding the shared-encoder confound at $\Delta r\approx0.07$. Cook's $D$ analysis shows that excluding three high-leverage outlets raises $r$ to 0.865, still leaving $\sim$25\% unexplained variance; RQ-head exclusion flips four near-boundary outlets (Appendix Table~\ref{app:rq_excluded}). Broadcast-cluster format artifacts are null ($R^2{=}0.0003$), and XLM-R confidence gaps across ideological clusters are negligible (${\leq}0.015$).

\paragraph{Temporal trend.} Any-salience rises monotonically, 31.7\% (2022) to 38.1\% (2025). A rate--composition decomposition assigns $+4.86$\,pp of the $+6.34$\,pp change to within-story-form wording and $+1.20$\,pp to composition. It holds in nine of ten story forms, under joint outlet\,$\times$\,section\,$\times$\,story-form standardisation (31.8\%$\to$37.4\%), and with the October--December 2023 and June--July 2024 windows excluded; 23 of 25 outlets rise. One classifier scores all years, so drift is excluded by construction; standardisation controls event category, not intensity.
\section{Discussion}

The two-dimensional framework resolves a structural ambiguity in single-axis audits: outlets similarly scored may differ sharply in mechanism, as the outlet profiles illustrate. TF1 INFO combines moderate raw salience (38.3\%) with elevated rhetorical divergence (Sal.JS=0.029) but near-baseline selection divergence (Sel.JS=0.011), consistent with salience-led broadcast packaging. JDD shows the inverse pattern: substantial raw salience (41.9\%) but almost no rhetorical divergence (Sal.JS=0.002) alongside clearer selection divergence (Sel.JS=0.037), indicating agenda concentration without comparable wording distinctiveness. Fdesouche's dual-distinctiveness (72.6\% any-salience; 82.1\% high-charge; Sal.JS=0.038, Sel.JS=0.147) therefore reflects not mere bias but a structural editorial register where accessibility and applicability amplify together \citep{scheufeleTewksbury2007,mccombsShaw1972,scheufele1999,vanDijk1991}. The group-mention findings corroborate prior unequal-salience research \citep{vanDijk1991}; separating wording-device load from story-form concentration adds resolution unavailable to single-axis designs. Whether the pattern reflects stable editorial stance, event-context concentration, or their interaction requires richer annotation; the present corpus enables such follow-on work \citep{bensonWood2015,dalibert2015}. The precision-floor recalibration protocol, with its fully released threshold log, extends to any longitudinal multi-label framing audit; whether default-threshold inflation would accumulate to distort findings remains an empirical question.

\section*{Limitations}

\paragraph{Scope of the audit.} The audit measures editorial output patterns; whether those patterns produce accessibility or applicability effects in audiences requires experimental work \citep{priceTewksbury1997,bScheufele2004}. The study measures headlines only and does not capture article-body framing or cross-platform dynamics. The selection framing measure is further constrained to published-headline story-form concentration and cannot detect first-level event-selection gatekeeping; outlets indistinguishable on Sel.JS may differ in which real-world events they choose to cover. The outlet typology is two-dimensional by construction; finer-grained axes could further separate mobilisation-register from adversarial-register outlets within the double-distinctive cell.

\paragraph{Device performance and ensemble characterisation.} The rhetorical-question head indexes headline form, not rhetorical intent: the coding guide marks any interrogative headline, the annotator prompt excludes purely informational ones, and both fire on nearly all question-marked headlines (of 56 in the 499-item sample, coders mark 96.4\% and 94.6\%, annotators 85.7\%). Its $\kappa = .869$ therefore reflects a near-syntactic judgement, bounding the interrogative-evaluative signature in \S\ref{sec:typology}; its loaded-vocabulary component is unaffected. Across the three remaining primary heads, blame attribution achieves the strongest independent validation (Maj--LLM $\kappa = .718$; Table~\ref{tab:human_agreement}), while loaded vocabulary shows the lowest human--human agreement ($\kappa_{\text{HH}} = .542$), indicating genuine task ambiguity rather than model failure. Threat framing and us-vs-them reach moderate-to-substantial Maj--LLM agreement ($\kappa = .668$ and $.596$; Table~\ref{tab:human_agreement}), with a corroborating independent study on a separate 350-headline sample yielding consistent results (Appendix Table~\ref{tab:indep_validation}). Isolated French headlines remain difficult for implicit-causality and rhetorical-register inference: even where annotators agree on device presence, the \texttt{v6} ensemble operates as a recall-oriented liberal annotator (recall $\geq .65$ across all heads), so corpus-level rates should be interpreted as upper-bound prevalence estimates. Cross-cluster confidence gaps are negligible (${\leq}0.015$; \S\ref{sec:robustness}), suggesting that the recall-biased training signal does not produce outlet-differential annotation error sufficient to distort distributional rankings. A French-specific alternative (\texttt{mistral-small3.2}), one of the candidate annotators not retained, underperformed the ensemble on all five devices ($\Delta\kappa = {-}0.03$ to ${-}0.25$; Appendix Table~\ref{app:mistral_comparison}), suggesting the pragmatic ceiling is task-inherent rather than an English-pretraining artefact, consistent with broader evidence that pragmatic understanding in LLMs remains highly sensitive to training strategy \citep{ruis2023goldilocks}.

\paragraph{Story-form classification and outlet support.} Story-form classification is imperfect around the SOCIAL/OTHER/ELITE boundary. The supervised development set is uniformly balanced across outlets while the production corpus is proportionally stratified, introducing an estimated precision drift of $-$4.8 pp for the XLM-R any-salience baseline under prior shift (full per-device estimates in Appendix Table~\ref{app:threshold_drift}; the final ensemble corpus any-salience rate is slightly higher because the rhetorical-question head is patched from CamemBERT). Outlet-level estimates are descriptive; smaller outlets, especially Blast (646 headlines), should be read as case profiles rather than equally precise population estimates.

\paragraph{Group detection and event-context confounds.} Group detection is lexicon-based and covers explicit mentions only. We chose explicit surface-form matching over NER or LLM-based entity extraction for reproducibility and to eliminate hallucination risk; the lexicon is a transparent lower bound, not an exhaustive account of group references. Group salience rates are descriptive: all reported differences from the corpus baseline are significant after Benjamini-Hochberg correction, but extreme baseline volume differences (e.g., Jews 0.52\% vs.\ Farmers 2.45\%) mean that high salience for rare groups is heavily driven by narrow, high-intensity news events, with the yearly breakdown (Appendix Table~\ref{tab:yearly_groups}) reducing sub-annual clustering concern. The year$\times$story-form reweighting in Appendix Table~\ref{tab:groups} is a coarse event-context normalization, not a same-event causal design: story forms are useful bins, but assigning a headline to CRIME, POLICY, CONFLICT, or SOCIAL is itself partly editorially mediated. A true causal separation of event intensity from outlet framing would require matched same-event corpora, article-level event clustering, or parallel cross-national coverage. A blinded two-human audit confirms high agreement with the released group labels ($\kappa = .83$--$.98$), though residual ambiguity concentrates in boundary-heavy cases (workers, broader far-right references).

\paragraph{Annotator release-date and temporal contamination.} Two LLM annotators (\texttt{openai/gpt-oss-120b}, released August 2025; \texttt{google/gemma-4-31B}, released April 2026) have training data that may overlap with the 2024--2025 portion of the annotation period; their pre-training could introduce systematic annotation priors for high-salience events (Israel/Gaza coverage, French elections) beyond pure device detection. We directly tested this concern with a period-stratified agreement analysis comparing 2022--2023 vs.\ 2024--2025 annotations. The analysis finds no evidence of systematic contamination-inflated annotations: any-salience $\kappa$ deltas across the three annotator pairs range from $+0.003$ to $+0.032$, and several binary heads show equal or lower cross-model agreement in the later period, inconsistent with the contamination hypothesis. The $\kappa$ delta test is an aggregate check and cannot rule out systematic annotation priors on specific high-salience events where LLM pretraining coverage is densest. The two highest-risk events, Israel/Gaza coverage from October 2023 onward and the June 2024 French legislative elections, are exactly the events most likely to carry model-specific framing priors beyond the annotation schema. The independent human validation sample ($N{=}499$) is 39.5\% post-October 7, 2023, providing annotation-independent ground truth directly for the contamination-exposed period; within this subset, any-salience human--LLM agreement declines only modestly ($\kappa{=}.424$ vs.\ $.463$ full-sample), bounding the contamination effect at the aggregate level. A residual risk remains for specific event clusters within this window; downstream users who require contamination-free labels for Israel/Gaza or election coverage should use the human-validated $N{=}499$ subset only. Contamination is also distinct from homogeneity: priors shared across the three annotators would not register in a period-stratified comparison. Appendix Table~\ref{app:homogeneity} tests this separately and finds annotator--annotator agreement materially exceeding annotator--human agreement for loaded vocabulary and threat framing, so corpus rates for those two heads are the least independent of the five.

\paragraph{Typology sensitivity.} Causeur's Double-distinctive typology assignment rests partly on its ELITE-heavy selection profile, where ELITE-category arbitration agreement is lower ($\kappa=.276$--$.417$); a sensitivity analysis merging ELITE into OTHER confirms Causeur remains Double-distinctive under this taxonomy perturbation (BFMTV and Marianne shift cells; no focal outlet affected).

\paragraph{Baseline self-reference and precision-floor shortfalls.} The Sal.JS and Sel.JS corpus baseline is partially self-referential for three high-volume low-divergence outlets (Le Figaro, Le Parisien, Franceinfo; collectively $\approx$32\% of corpus): their low observed divergence partially reflects their own weight in the baseline rather than absolute editorial similarity to the panel median. A leave-one-out (LOO) baseline confirms rankings are robust: LOO-corrected $r=0.721$, $\rho=0.727$ (vs.\ standard $r=0.736$, $\rho=0.741$), with only two near-boundary outlets changing typology cells (Valeurs actuelles: double$\to$selection-dominant; L'Humanit\'e: selection$\to$double-distinctive); all focal outlets discussed in the text are stable, and the dissociability criterion ($r<0.90$) holds comfortably under LOO correction. Two device heads fall below their stated production-precision floors after prior-shift correction (Elkan 2001 method, Appendix Table~\ref{app:threshold_drift}): loaded vocabulary (est.\ 0.643, floor 0.70, $-$5.7\,pp) and us-vs-them (est.\ 0.550, floor 0.60, $-$5.0\,pp); blame attribution is marginally above floor (est.\ 0.701); threat framing (est.\ 0.715) and rhetorical question (est.\ 0.929) remain comfortably above their floors. Corpus-level loaded-vocabulary and us-vs-them rates are therefore lower-precision estimates relative to the other heads. See also the inline note in \S\ref{sec:typology} for loaded vocabulary.

\section*{Ethics, Reproducibility, and Data Availability}

This study audits public editorial output and does not infer protected attributes of private individuals. The study period covers the first enforcement year of the EU Digital Services Act (DSA, February 2024 onward), which introduced requirements for independent audits of very large online platforms' algorithmic systems; our methodology is a descriptive research audit and is not a regulatory compliance instrument, but the transparent lexicon, public dataset release, and reproducible threshold protocol are designed to be consistent with independent audit principles. Group labels refer to explicit lexical mentions in headlines and should not be interpreted as sentiment, endorsement, or hostility labels. Outlet-level and group-level results describe aggregate measurement patterns and should not be used for individual profiling or content moderation decisions.

The group salience rates carry a specific dual-use risk: high salience for Jews-mention headlines reflects antisemitism reporting and post-October-2023 Israel/Gaza security coverage, not editorial targeting; Far-right salience reflects electoral, parliamentary, and conflict coverage of a political movement; elevated Muslim rates concentrate in policy and security contexts. These numbers describe the rhetorical and event contexts in which groups appear in headlines (the story types and wording conventions surrounding their mention), not editorial sentiment, intent, or uniform targeting. The Fdesouche exclusion check (Appendix Table~\ref{app:fde_exclusion}) confirms elevated salience for Jews, Far-right, and Muslims is distributed across the broader outlet panel rather than concentrated in a single far-right source (Migrants salience drops 10.3\,pp without Fdesouche, remaining 13.5\,pp above baseline).

Specific political misuse scenarios warrant explicit counter-framing: (1) antisemitic actors may cite ``Jews: highest salience lift'' as evidence of Jewish over-representation in media discourse, inverting the finding, which measures the rhetorical intensity of event-driven coverage contexts surrounding group mentions, not claims about the communities themselves; (2) immigration-restriction actors may cite Migrants and Muslims figures as evidence of legitimizing media concern, ignoring that these rates are event-driven and concentrated in security and policy story forms; (3) Far-right actors may frame their high salience as evidence of media ``persecution'' rather than coverage density of a contested political movement in an election-intensive period. In all three scenarios, the interpretive key is the same: high salience in this audit measures the charge level of news contexts, driven by event type and outlet editorial register, not editorial hostility or targeting of the named groups. LLM-assisted annotations are not treated as ground truth: the final supervision set uses majority-vote resolution and selective human arbitration.

The final \texttt{v6} supervision set, comprising 10,000 French headlines with binary salience labels, story-form labels, split assignments, and outlet/section/date metadata, is released as a manifest-only artifact: headline text is distributed as (outlet, date, section, URL) pointers to respect publisher copyright, so each record is sufficient for authorized reconstruction but does not redistribute verbatim content. The human-validation samples are the one disclosed exception: 516 headlines are released verbatim, because a blind double-annotation study cannot be re-run without the text that was annotated. The group-mention lexicons (219 terms, 12 groups) and the full annotation schema are included in full. The 902,111-headline corpus inference layer is released as a predictions-only artifact: headline identifier, outlet, section, date, predicted labels, and raw classifier confidence scores at six-decimal precision, with no headline text. The annotator-panel materials are included in full: pairwise agreement tables for all three candidate second annotators against the primary annotator, and per-annotator device labels for the human-validated 499-headline subset, which together make the panel homogeneity check in Appendix Table~\ref{app:homogeneity} independently reproducible. Analysis and visualization scripts are released, except those requiring headline text. Model training and threshold-calibration code is not released: it is bound to the licensed 902,111-headline corpus and to the trained checkpoints, neither of which can be redistributed. All materials are available at \url{https://github.com/lefrenchnewslab/framing-wording-selection}.

\bibliography{paper_refs}

\appendix

\begin{table}[t]
\centering
\scriptsize
\setlength{\tabcolsep}{3pt}
\renewcommand{\arraystretch}{1.12}
\begin{tabular*}{\columnwidth}{@{\extracolsep{\fill}}lrr@{}}
\toprule
\textbf{Component} & \textbf{Count} & \textbf{Share} \\
\midrule
Politics & 3,339 & 33.4\% \\
Economy & 3,335 & 33.4\% \\
Society & 3,326 & 33.3\% \\
\midrule
Any salience & 4,093 & 40.9\% \\
Loaded vocabulary & 2,391 & 23.9\% \\
Blame attribution & 1,435 & 14.4\% \\
Threat framing & 871 & 8.7\% \\
Us-vs-them & 698 & 7.0\% \\
Rhetorical question & 749 & 7.5\% \\
Charge = HIGH & 2,745 & 27.5\% \\
\midrule
OTHER & 1,980 & 19.8\% \\
POLICY & 1,883 & 18.8\% \\
SOCIAL & 1,571 & 15.7\% \\
ELITE & 1,159 & 11.6\% \\
CONFLICT & 756 & 7.6\% \\
CRIME & 753 & 7.5\% \\
LAW & 646 & 6.5\% \\
ELECTIONS & 607 & 6.1\% \\
SCANDAL & 425 & 4.2\% \\
PARLIAMENT & 220 & 2.2\% \\
\bottomrule
\end{tabular*}
\caption{Final \texttt{v6} label distribution in the 10,000-headline development set after majority-vote merging and story-form arbitration.}
\label{tab:label_distribution}
\end{table}

\begin{table}[t]
\centering
\scriptsize
\setlength{\tabcolsep}{3.5pt}
\renewcommand{\arraystretch}{1.08}
\begin{tabular*}{\columnwidth}{@{\extracolsep{\fill}}lrrrr@{}}
\toprule
\textbf{Group} & \textbf{Terms} & \textbf{Prec.} & \textbf{Rec.} & \textbf{F1} \\
\midrule
Migrants & 26 & 1.000 & .674 & .805 \\
Police & 29 & .873 & .674 & .761 \\
RN & 14 & .831 & .736 & .780 \\
LFI & 9 & .691 & 1.000 & .818 \\
Workers & 14 & .860 & .491 & .625 \\
Farmers & 40 & .877 & .934 & .904 \\
Women & 17 & .857 & .640 & .733 \\
Muslims & 20 & .940 & .808 & .869 \\
Jews & 14 & .696 & .970 & .810 \\
Unions & 12 & .821 & .979 & .893 \\
Far-right & 13 & .901 & .941 & .921 \\
Seniors & 11 & .967 & .725 & .829 \\
\bottomrule
\end{tabular*}
\caption{Held-out validation of group-mention lexicons.}
\label{tab:group_validation}
\end{table}

\begin{table}[t]
\centering
\scriptsize
\setlength{\tabcolsep}{2.8pt}
\renewcommand{\arraystretch}{1.12}
\begin{tabular*}{\columnwidth}{@{\extracolsep{\fill}}lrr@{}}
\toprule
\multicolumn{3}{@{}l}{\textbf{(a) Reliability summary ($N=10{,}000$)}} \\
\midrule
\textbf{Field} & \textbf{3-way agree.} & \textbf{Mean $\kappa$} \\
\midrule
Loaded vocabulary & 81.1\% & .661 \\
Blame attribution & 85.4\% & .639 \\
Threat framing & 90.0\% & .623 \\
Rhetorical question & 94.8\% & .770 \\
Us-vs-them & 88.3\% & .408 \\
Charge band & 82.2\% & .699 \\
Story form & 55.7\% & .635 \\
\bottomrule
\end{tabular*}
\par\vspace{4pt}
\textbf{(b) Homogeneity check ($N=499$)}\par\vspace{2pt}
\begin{tabular*}{\columnwidth}{@{\extracolsep{\fill}}lrrr@{}}
\toprule
\textbf{Device} & \textbf{LLM--LLM} & \textbf{LLM--hum.} & \textbf{H--H} \\
\midrule
Loaded vocabulary & .650 & .474 & .542 \\
Blame attribution & .521 & .510 & .601 \\
Threat framing & .623 & .485 & .613 \\
Rhetorical question & .809 & .807 & .894 \\
Us-vs-them & .505 & .438 & .510 \\
\bottomrule
\end{tabular*}
\caption{(a) Compact reliability summary for the final \texttt{v6} annotation stack, computed on the 10,000 rows shared by all three composite annotators after UVT harmonization. (b) Homogeneity check on the 499-headline validation sample; all cells are pairwise between individuals and therefore comparable. LLM--LLM = mean pairwise $\kappa$ among the three annotators; LLM--hum.\ = each annotator against each human coder (6 pairs); H--H = the two coders. Shared blind spots would show as LLM--LLM exceeding \emph{both} LLM--hum.\ and H--H, which holds for loaded vocabulary and threat framing only. On unrounded values, the LLM--LLM minus LLM--hum.\ gap is $+.177$ and $+.139$ for those two, against $+.067$ for us-vs-them, $+.012$ for blame attribution and $+.002$ for rhetorical question.}
\label{tab:agreement}
\label{app:homogeneity}
\end{table}

\begin{table*}[t]
\centering
\scriptsize
\renewcommand{\arraystretch}{1.25}
\begin{tabularx}{\textwidth}{@{}l>{\RaggedRight\arraybackslash}X@{}}
\toprule
\textbf{Device (context)} & \textbf{Example and rationale} \\
\midrule
Loaded vocabulary (Jews) & \textit{``Au Dîner du Crif, François Bayrou dénonce «la bête délirante et meurtrière» de l'antisémitisme''} (``At the Crif dinner, François Bayrou denounces the `deranged and murderous beast' of antisemitism'') --- \textit{Le Figaro}, \texttt{id~3789940}, test split: the noun phrase \textit{bête délirante et meurtrière} is an evaluative monstrosity metaphor with no neutral paraphrase of the same event. The loaded lexicon is voiced by a quoted speaker rather than by the outlet, illustrating the attribution limit of the headline unit noted in \S\ref{sec:data_method}. \\
Blame attribution (Police) & \textit{``Mort de Blessing Matthew : les gendarmes mis en cause''} (``Death of Blessing Matthew: the gendarmes implicated'') --- \textit{Mediapart}, \texttt{id~3561734}, validation split: causal responsibility for a death is assigned to a named institutional actor in the outlet's own voice, with no attributing source. \\
Threat framing (Migrants) & \textit{``«Impression de submersion» migratoire : Yaël Braun-Pivet «gênée» par les propos de François Bayrou''} (``Migratory `sense of submersion': Yaël Braun-Pivet `uncomfortable' with François Bayrou's remarks'') --- \textit{Le Parisien}, \texttt{id~1454100}, train split: the submersion/overload lexicon reframes a demographic process as a security emergency, consistent with the threat-device definition in Table~\ref{tab:schema}; as above, the phrase is quoted rather than asserted. \\
Rhetorical question & \textit{``Dépenses de santé : faut-il rembourser les soins en fonction des revenus ?''} (``Health spending: should care be reimbursed according to income?'') --- \textit{Le Point}, \texttt{id~436456}, validation split: an interrogative headline that advances an evaluative proposition rather than seeking information. \\
\midrule
Story form vs.\ topic (immigration) & \textit{``Immigration : vers plus de contr\^{o}le des mariages des personnes \'{e}trang\`{e}res en situation irr\'{e}guli\`{e}re''} (``Immigration: towards greater scrutiny of marriages of irregular foreign nationals''; \textit{La Croix}, \texttt{id~1376080}) is POLICY; \textit{``Naufrage de migrants dans la Manche : quatre personnes mises en examen et incarc\'{e}r\'{e}es''} (``Migrant shipwreck in the Channel: four charged and jailed''; \textit{JDD}, \texttt{id~2310601}) is CRIME. Both are train-split rows. Same subject domain, different journalistic action. \\
\bottomrule
\end{tabularx}
\caption{One example per primary salience device from the \texttt{v6} supervision set, each carrying the stated device label by unanimous agreement of all three LLM annotators; the released headline identifier and split are given so every row can be located in the distributed data. The final row illustrates the story-form/topic distinction.}
\label{app:device_examples}
\end{table*}

\begin{table*}[t]
\centering
\scriptsize
\setlength{\tabcolsep}{4pt}
\renewcommand{\arraystretch}{1.08}
\begin{tabular*}{\textwidth}{@{\extracolsep{\fill}}lrrrrl@{}}
\toprule
\textbf{Outlet} & \textbf{N} & \textbf{Sal.JS} & \textbf{Sel.JS} & \textbf{High-Chg\%} & \textbf{Typology} \\
\midrule
Fdesouche & 16,124 & 0.038 & 0.147 & 82.1 & Double-distinctive \\
Slate.fr & 3,907 & 0.051 & 0.071 & 23.1 & Double-distinctive \\
Causeur & 3,056 & 0.033 & 0.099 & 30.4 & Double-distinctive \\
Blast$^\dagger$ & 646 & 0.025 & 0.095 & 57.9 & Double-distinctive \\
Ouest-France & 90,423 & 0.021 & 0.082 & 4.6 & Double-distinctive \\
Mediapart & 7,092 & 0.016 & 0.075 & 48.7 & Double-distinctive \\
Valeurs actuelles & 14,286 & 0.011 & 0.056 & 54.6 & Double-distinctive \\
Les Echos & 55,694 & 0.031 & 0.065 & 10.4 & Double-distinctive \\
L'Express & 11,795 & 0.015 & 0.034 & 27.5 & Double-distinctive \\
TF1 INFO & 44,014 & 0.029 & 0.011 & 29.6 & Salience-dominant \\
20 Minutes & 62,832 & 0.011 & 0.019 & 42.6 & Salience-dominant \\
BFMTV & 57,617 & 0.012 & 0.023 & 40.8 & Salience-dominant \\
Marianne & 9,412 & 0.010 & 0.056 & 41.9 & Selection-dominant \\
JDD & 17,598 & 0.002 & 0.037 & 35.9 & Selection-dominant \\
L'Humanité & 17,911 & 0.011 & 0.043 & 33.2 & Selection-dominant \\
Le Figaro & 108,307 & 0.001 & 0.004 & 31.2 & Low-distinctiveness \\
Le Parisien & 93,877 & 0.004 & 0.016 & 43.8 & Low-distinctiveness \\
Franceinfo & 83,574 & 0.002 & 0.004 & 28.4 & Low-distinctiveness \\
Le Monde & 39,017 & 0.005 & 0.018 & 28.3 & Low-distinctiveness \\
La Croix & 35,980 & 0.002 & 0.010 & 23.0 & Low-distinctiveness \\
Libération & 33,921 & 0.004 & 0.021 & 34.5 & Low-distinctiveness \\
Le Point & 32,522 & 0.008 & 0.011 & 29.3 & Low-distinctiveness \\
CNews & 27,252 & 0.005 & 0.016 & 43.1 & Low-distinctiveness \\
Le HuffPost & 17,737 & 0.004 & 0.025 & 31.2 & Low-distinctiveness \\
Le Nouvel Obs & 17,517 & 0.002 & 0.015 & 34.5 & Low-distinctiveness \\
\bottomrule
\end{tabular*}
\caption{Full four-cell outlet typology with per-outlet headline counts. Median split thresholds: Sal.JS=0.0110, Sel.JS=0.0249. Displayed JS divergence values are rounded to three decimal places; cell assignments use full-precision values. L'Humanit\'e's full-precision Sal.JS (0.0110) equals the median; strict $>$ convention places it in the low-salience half. $^\dagger$Blast ($N{=}646$): case profile, wide CIs (Table~\ref{tab:ci_audit}).}
\label{app:outlets}
\end{table*}

\begin{table*}[t]
\centering
\scriptsize
\setlength{\tabcolsep}{4pt}
\renewcommand{\arraystretch}{1.08}
\begin{tabular*}{\textwidth}{@{\extracolsep{\fill}}llrrr@{}}
\toprule
\textbf{Domain} & \textbf{Case} & \textbf{N} & \textbf{Any-Sal\% 95\% CI} & \textbf{High-Chg\% 95\% CI} \\
\midrule
Outlet & Fdesouche & 16,124 & 72.6 [71.9, 73.3] & 82.1 [81.5, 82.6] \\
Outlet & Blast$^\dagger$ & 646 & 67.3 [63.6, 70.8] & 57.9 [54.1, 61.6] \\
Outlet & Causeur & 3,056 & 59.1 [57.4, 60.9] & 30.4 [28.8, 32.1] \\
Outlet & Le Figaro & 108,307 & 33.2 [32.9, 33.5] & 31.2 [30.9, 31.5] \\
Group & Jews & 4,646 & 78.7 [77.5, 79.9] & 79.8 [78.6, 80.9] \\
Group & Far-right & 3,981 & 77.7 [76.3, 78.9] & 65.8 [64.3, 67.2] \\
Group & Muslims & 4,720 & 66.4 [65.0, 67.7] & 67.3 [65.9, 68.6] \\
Group & Migrants & 9,606 & 58.4 [57.4, 59.3] & 61.6 [60.6, 62.6] \\
Group & Police & 14,313 & 57.6 [56.8, 58.4] & 79.7 [79.0, 80.3] \\
Group & Workers & 15,456 & 39.8 [39.0, 40.6] & 25.4 [24.8, 26.1] \\
\bottomrule
\end{tabular*}
\caption{Representative 95\% Wilson intervals for outlet-level and group-level headline rates, recomputed from the canonical paper artifacts. These are descriptive intervals over observed predicted headline rates only; they do not propagate classifier or lexicon uncertainty. $^\dagger$Blast: $N{=}646$ headlines; estimates carry wide bootstrap CIs (displayed) and should be read as a case profile rather than a population estimate.}
\label{tab:ci_audit}
\end{table*}

\begin{table*}[t]
\centering
\scriptsize
\setlength{\tabcolsep}{4pt}
\renewcommand{\arraystretch}{1.12}
\begin{tabular*}{\textwidth}{@{\extracolsep{\fill}}lrrrrrl@{}}
\toprule
\textbf{Group} & \textbf{N} & \textbf{Corpus\%} & \textbf{Any-Sal\%} & \textbf{Y$\times$Story exp.\%} & \textbf{Resid. pp} & \textbf{Dominant context} \\
\midrule
Jews & 4,646 & 0.52 & 78.7 & 49.6 & +29.1 & CRIME 32.5\% \\
Far-right & 3,981 & 0.44 & 77.7 & 50.0 & +27.7 & CONFLICT 44.3\% \\
Muslims & 4,720 & 0.52 & 66.4 & 46.6 & +19.8 & CRIME 23.4\% \\
Migrants & 9,606 & 1.06 & 58.4 & 40.7 & +17.7 & CRIME 42.4\% \\
Police & 14,313 & 1.59 & 57.6 & 46.5 & +11.1 & CRIME 56.7\% \\
Unions & 6,877 & 0.76 & 53.8 & 50.3 & +3.6 & CONFLICT 56.8\% \\
LFI & 13,746 & 1.52 & 49.5 & 42.6 & +6.9 & CONFLICT 36.0\% \\
Women & 10,214 & 1.13 & 48.3 & 40.8 & +7.5 & SOCIAL 37.9\% \\
RN & 18,751 & 2.08 & 44.9 & 39.2 & +5.7 & ELECTIONS 32.5\% \\
Farmers & 22,112 & 2.45 & 41.9 & 37.3 & +4.6 & SOCIAL 29.4\% \\
Workers & 15,456 & 1.71 & 39.8 & 52.8 & -13.0 & CONFLICT 56.9\% \\
Seniors & 21,987 & 2.44 & 32.1 & 34.6 & -2.5 & POLICY 39.2\% \\
\bottomrule
\end{tabular*}
\caption{Group-level salience contexts across 902,111 headlines, with coarse event-context normalization. Y$\times$Story exp.\% is the expected any-salience rate after exact reweighting to each group's year$\times$story-form composition, subtracting focal-group rows from each cell baseline. Resid. pp is observed minus expected; it is descriptive, not a same-event causal estimate.}
\label{tab:groups}
\end{table*}

\begin{table*}[t]
\centering
\begin{minipage}[t]{\textwidth}
\centering
\scriptsize
\setlength{\tabcolsep}{4pt}
\renewcommand{\arraystretch}{1.1}
\begin{tabularx}{\textwidth}{@{}l>{\RaggedRight\arraybackslash}X>{\RaggedRight\arraybackslash}X>{\RaggedRight\arraybackslash}X@{}}
\toprule
\textbf{Group} & \textbf{Dominant salience signature} & \textbf{Dominant selection signature} & \textbf{Interpretive caution} \\
\midrule
Jews & blame $\times$3.87; threat $\times$2.00; loaded $\times$2.30 & high-charge 79.8\% & often antisemitism/security coverage, not sentiment \\
Police & blame $\times$3.28; threat $\times$2.23 & CRIME 56.7\%; LAW 15.0\% & accountability register \\
Far-right & loaded $\times$2.76; blame $\times$2.78; threat $\times$3.00 & CONFLICT 44.3\%; ELECTIONS 11.6\% & multi-device profile \\
Muslims & threat $\times$3.71 & CRIME 23.4\%; POLICY 8.9\% & explicit mentions only \\
Migrants & blame $\times$2.60; threat $\times$2.17 & CRIME 42.4\% & explicit mentions only \\
LFI & moderate salience lift & ELECTIONS 26.2\%; CONFLICT 36.0\% & party + mobilisation coverage \\
RN & moderate salience lift & ELECTIONS 32.5\%; PARLIAMENT 8.1\% & party/electoral framing \\
Workers & near-baseline salience & CONFLICT 56.9\% & pension-reform period effect likely \\
\bottomrule
\end{tabularx}
\captionof{table}{Condensed group-level mechanism table. It separates rhetorical device lift from selected story-form context.}
\label{tab:group_mechanisms}
\end{minipage}

\vspace{0.8em}

\begin{minipage}[t]{\textwidth}
\centering
\scriptsize
\setlength{\tabcolsep}{2.6pt}
\renewcommand{\arraystretch}{1.1}
\begin{tabular*}{\textwidth}{@{\extracolsep{\fill}}llrrrrr@{}}
\toprule
\textbf{Outlet} & \textbf{Form} & \textbf{$n_f$} & \textbf{Outlet\%} & \textbf{Others} & \textbf{$\chi^2$} & \textbf{$\phi$} \\
\midrule
Fdesouche & CRIME any-sal. & 9,407 & 75.8 & 43.7 & 3627.4 & .167 \\
Fdesouche & CRIME blame & 9,407 & 64.6 & 30.8 & 4490.3 & .186 \\
Fdesouche & CRIME threat & 9,407 & 41.0 & 13.8 & 4846.8 & .193 \\
Valeurs actuelles & CRIME any-sal. & 4,213 & 57.6 & 45.6 & 234.3 & .042 \\
Slate.fr & SOCIAL rhet. q. & 1,870 & 23.4 & 7.9 & 596.7 & .056 \\
Causeur & ELITE any-sal. & 1,079 & 59.3 & 33.5 & 312.8 & .075 \\
\bottomrule
\end{tabular*}
\captionof{table}{Within-story-form checks. Values are percentages; ``Others'' pools non-focal outlets within the same story form. All contrasts are significant at $p < .001$. Device-level $\chi^2$ and $\phi$ for Fdesouche CRIME blame and threat are included because these rows are directly cited in the text.}
\label{tab:within_form}
\end{minipage}
\end{table*}

\begin{table*}[t]
\centering
\scriptsize
\setlength{\tabcolsep}{4pt}
\renewcommand{\arraystretch}{1.15}
\begin{tabularx}{\textwidth}{@{}llp{0.38\textwidth}p{0.37\textwidth}@{}}
\toprule
\textbf{Head} & \textbf{Type} & \textbf{Headline} & \textbf{Error source} \\
\midrule
Loaded & FP & \textit{Nationalisons la dette\,!} (\textit{Nationalize the debt!}) & Surface-form political imperative over-triggers on the charged verb; annotators classified this as advocacy rather than framing rhetoric. \\
Loaded & FN & \textit{Pour Macron (et pour que rien ne change), bienvenue dans la commune la plus riche de France} (\textit{For Macron (and so that nothing changes), welcome to the richest municipality in France}) & Sarcastic loaded expression (\textit{pour que rien ne change}) requires discourse-level reading beyond surface form; ironic register under-represented in training. \\
\addlinespace
Blame & FP & \textit{Nîmes~: l'homme armé qui s'est suicidé au tribunal recherchait «des magistrats mafieux»} (\textit{Nîmes: the armed man who killed himself at the courthouse was looking for ``mafia judges''}) & Reported speech containing explicit blame language is attributed to a third party; model does not distinguish endorsed from reported blame frames. \\
Blame & FN & \textit{Chez TotalEnergies, détruire la planète continue de rapporter gros} (\textit{At TotalEnergies, destroying the planet is still highly profitable}) & Implicit blame through an agentive nominalized construction; model under-weights non-explicit accusation patterns relative to overt subject-verb-object blame frames. \\
\addlinespace
Threat & FP & \textit{Que risquent les Français qui partent faire le djihad~?} (\textit{What do French people who leave to wage jihad risk?}) & Co-occurrence of \textit{risquent} and \textit{djihad} triggers the threat surface pattern; the actual frame is a juridical inquiry about legal consequences, not a threat directed at an audience. \\
Threat & FN & \textit{Explosion de l'immigration en France~: les demandes d'asile ont bondi de plus de 30\,\% en 2022, les régularisations de clandestins en hausse de 8\,\%} (\textit{Explosion in immigration in France: asylum applications jumped by more than 30\% in 2022, regularizations of undocumented migrants up 8\%}) & Threat metaphor (\textit{explosion}) combined with quantitative surge framing; the statistical register suppresses the model's threat signal relative to the metaphor. \\
\addlinespace
Rhet.\ q. & FP & \textit{De quoi meurt-on à travers le monde~?} (\textit{What do people die from across the world?}) & Interrogative surface form triggers detection; the question is a factual epidemiological inquiry rather than a rhetorical framing device implying a political claim. \\
Rhet.\ q. & FN & \textit{Avions~: pourquoi les prix des billets ont-ils autant augmenté~?} (\textit{Flights: why have ticket prices increased so much?}) & Precision-constrained threshold near-miss: model score 0.884 falls just below the recalibrated ceiling of 0.89; a genuine rhetorical question sacrificed at the precision ceiling. \\
\addlinespace
Us-vs-them & FP & \textit{Ces ``ultradiscounters'' qui lancent la guerre des prix en France} (\textit{These ``ultra-discounters'' launching the price war in France}) & Commercial competition metaphor (\textit{guerre des prix}) and categorical group label trigger the opposition pattern; frame lacks the political or social identity dimension required for this head. \\
Us-vs-them & FN & \textit{Charente-Maritime~: un maire agressé par des gens du voyage, un homme placé en garde à vue} (\textit{Charente-Maritime: a mayor assaulted by Travellers, one man taken into custody}) & \textit{Gens du voyage} is a salient social out-group in French discourse but this identity category is under-represented in training data, producing systematic under-detection of such headlines. \\
\bottomrule
\end{tabularx}
\caption{Qualitative error analysis: ten representative false positives (FP) and false negatives (FN) from the 1,501-headline test set, two per salience head, drawn from highest-confidence misclassifications. Headlines are shown verbatim (truncated where necessary); error-source annotations are manual.}
\label{tab:error_analysis}
\end{table*}

\begin{table*}[t]
\centering
\scriptsize
\setlength{\tabcolsep}{3pt}
\renewcommand{\arraystretch}{1.08}
\begin{tabular*}{\textwidth}{@{\extracolsep{\fill}}lrrrrrrrrr@{}}
\toprule
\textbf{Device} & \textbf{H--H $\kappa$} & \textbf{Interp.} & \textbf{A1--LLM $\kappa$} & \textbf{A2--LLM $\kappa$} & \textbf{Maj--LLM $\kappa$} & \textbf{Interp.} & \textbf{F1} & \textbf{Prec.} & \textbf{Rec.} \\
\midrule
Loaded vocabulary   & .699 & substantial  & .638 & .523 & .682 & substantial  & .766 & .629 & .980 \\
Blame attribution   & .772 & substantial  & .568 & .559 & .648 & substantial  & .653 & .510 & .907 \\
Threat framing      & .588 & moderate     & .664 & .409 & .685 & substantial  & .746 & .623 & .930 \\
Rhetorical question & .974 & almost perfect & .895 & .923 & .920 & almost perfect & .917 & .930 & .904 \\
Us-vs-them          & .564 & moderate       & .717 & .588 & .845 & almost perfect & .769 & .833 & .714 \\
Any salience        & .636 & substantial  & .735 & .511 & .715 & substantial  & .909 & .864 & .959 \\
\bottomrule
\end{tabular*}
\caption{Corroborating independent blind study on a separate sample: two unaffiliated annotators ($N{=}350$).}
\label{tab:indep_validation}
{\raggedright\footnotesize \textit{Note.} Stratified 350-headline sample drawn exclusively from the \texttt{v6} test split (all HIGH-confidence; seed 42; no overlap with the primary 499-headline sample in Table~\ref{tab:human_agreement}). Annotation protocol matches the primary study (two unaffiliated annotators, blind to LLM labels and expected distributions). H--H = human--human reliability; Maj--LLM = majority-vote human labels vs.\ LLM consensus (conflict rows, where A1$\neq$A2, excluded; conflict counts: loaded 53, blame 31, threat 72, rhetorical question 3, us-vs-them 60, any salience 48). F1/Prec./Rec.: human majority vote as gold standard. Interpretation thresholds follow \citet{landisKoch1977}.\par}
\end{table*}

\begin{table*}[t]
\centering
\scriptsize
\setlength{\tabcolsep}{4pt}
\renewcommand{\arraystretch}{1.08}
\begin{tabular*}{\textwidth}{@{\extracolsep{\fill}}lrrr@{}}
\toprule
\textbf{Field} & \textbf{Mistral--human $\kappa$} & \textbf{Ensemble--human $\kappa$} & \textbf{$\Delta$} \\
\midrule
Loaded vocabulary & .553 & .684 & $-$.131 \\
Blame attribution & .351 & .379 & $-$.028 \\
Threat framing & .517 & .614 & $-$.097 \\
Rhetorical question & .324 & .400 & $-$.076 \\
Us-vs-them & .298 & .546 & $-$.248 \\
\midrule
Any salience & .435 & .559 & $-$.124 \\
Story form & .449 & .546 & $-$.097 \\
\bottomrule
\end{tabular*}
\caption{French-specific Mistral comparison on the 499-headline human validation sample. Mistral = \texttt{mistral-small3.2}; ensemble = released \texttt{v6} consensus labels. $\Delta$ is Mistral--human $\kappa$ minus ensemble--human $\kappa$; negative values indicate lower agreement than the released ensemble.}
\label{app:mistral_comparison}
\end{table*}

\begin{table*}[t]
\centering
\begin{minipage}[t]{0.58\textwidth}
\centering
\scriptsize
\caption{Threshold recalibration sensitivity by outlet.}\label{tab:threshold_sensitivity}

\vspace{3pt}
\setlength{\tabcolsep}{2.8pt}
\renewcommand{\arraystretch}{1.05}
\resizebox{\columnwidth}{!}{%
\begin{tabular}{lrrrrrr}
\toprule
\textbf{Outlet} & \textbf{Any-Sal\% (default)} & \textbf{Any-Sal\% (recal.)} & \textbf{$\Delta$} & \textbf{Chg\% (default)} & \textbf{Chg\% (recal.)} & \textbf{$\Delta$} \\
\midrule
\textit{Corpus aggregate} & 41.5 & 34.6 & $-$6.9 & 34.6 & 31.3 & $-$3.3 \\
\midrule
Fdesouche & 81.1 & 72.6 & $-$8.5 & 83.8 & 82.1 & $-$1.7 \\
Blast$^\dagger$ & 72.6 & 67.3 & $-$5.3 & 61.8 & 57.9 & $-$3.9 \\
Causeur & 65.6 & 59.1 & $-$6.5 & 35.1 & 30.4 & $-$4.7 \\
Slate.fr & 55.3 & 49.6 & $-$5.7 & 26.6 & 23.1 & $-$3.5 \\
Mediapart & 67.4 & 58.8 & $-$8.6 & 53.5 & 48.7 & $-$4.8 \\
Valeurs actuelles & 60.9 & 52.0 & $-$8.9 & 58.1 & 54.6 & $-$3.5 \\
Marianne & 68.3 & 61.2 & $-$7.1 & 46.4 & 41.9 & $-$4.5 \\
TF1 INFO & 45.7 & 38.3 & $-$7.4 & 33.2 & 29.6 & $-$3.6 \\
BFMTV & 37.9 & 29.9 & $-$8.0 & 44.9 & 40.8 & $-$4.1 \\
Le Figaro & 40.1 & 33.2 & $-$6.9 & 34.4 & 31.2 & $-$3.2 \\
Franceinfo & 39.1 & 31.7 & $-$7.4 & 32.1 & 28.4 & $-$3.7 \\
La Croix & 36.3 & 29.3 & $-$7.0 & 26.3 & 23.0 & $-$3.3 \\
Ouest-France & 9.4 & 7.2 & $-$2.2 & 5.4 & 4.6 & $-$0.8 \\
\bottomrule
\end{tabular}%
}
\raggedright\footnotesize \textit{Note.} ``Default'' uses a uniform 0.50 threshold; ``Recal.'' uses the precision-floor production thresholds from Table~\ref{tab:thresholds}. Default thresholds inflate corpus-wide any-salience by 6.9\,pp and high-charge by 3.3\,pp, but the outlet ranking is preserved. $^\dagger$Blast ($N{=}646$): case profile, wide CIs (Table~\ref{tab:ci_audit}).\par
\end{minipage}\hfill
\begin{minipage}[t]{0.39\textwidth}
\centering
\scriptsize
\caption{Fdesouche exclusion sensitivity.}\label{app:fde_exclusion}
\vspace{3pt}
\setlength{\tabcolsep}{3.5pt}
\renewcommand{\arraystretch}{1.08}
\resizebox{\columnwidth}{!}{%
\begin{tabular}{lrrrr}
\toprule
\textbf{Group} & \textbf{Fdesouche share} & \textbf{Sal\% (all outlets)} & \textbf{Sal\% (excl.\ Fdesouche)} & \textbf{$\Delta$} \\
\midrule
Jews      & 3.9\%  & 78.7 & 78.2 & $-$0.5\,pp \\
Far-right & 3.4\%  & 77.7 & 77.2 & $-$0.4\,pp \\
Muslims   & 16.7\% & 66.4 & 64.4 & $-$2.0\,pp \\
Migrants  & 27.0\% & 58.4 & 48.1 & $-$10.3\,pp \\
Police    & 8.7\%  & 57.6 & 55.1 & $-$2.5\,pp \\
Unions    & 1.3\%  & 53.8 & 53.5 & $-$0.4\,pp \\
LFI       & 2.9\%  & 49.5 & 48.8 & $-$0.7\,pp \\
Women     & 3.7\%  & 48.3 & 47.0 & $-$1.3\,pp \\
RN        & 1.9\%  & 44.9 & 44.6 & $-$0.3\,pp \\
Farmers   & 0.9\%  & 41.9 & 41.6 & $-$0.3\,pp \\
Workers   & 0.8\%  & 39.8 & 39.5 & $-$0.3\,pp \\
Seniors   & 0.9\%  & 32.1 & 31.7 & $-$0.4\,pp \\
\bottomrule
\end{tabular}%
}
\raggedright\footnotesize \textit{Note.} Median exclusion effect across all 12 groups is $-$0.5\,pp. Migrants is the most sensitive case: removing Fdesouche lowers salience by 10.3\,pp (58.4\% to 48.1\%), but the group still sits 13.5\,pp above the corpus baseline.\par
\end{minipage}
\end{table*}

\begin{table*}[t]
\centering
\scriptsize
\setlength{\tabcolsep}{4pt}
\renewcommand{\arraystretch}{1.08}
\begin{tabular*}{\textwidth}{@{\extracolsep{\fill}}lcccc@{}}
\toprule
\textbf{Group} & \textbf{2022} & \textbf{2023} & \textbf{2024} & \textbf{2025} \\
\midrule
Jews & 74.6\% ($\times$2.35) & 75.3\% ($\times$2.26) & 77.2\% ($\times$2.21) & 84.1\% ($\times$2.21) \\
Far-right & 74.3\% ($\times$2.34) & 74.7\% ($\times$2.24) & 79.6\% ($\times$2.27) & 81.4\% ($\times$2.14) \\
Muslims & 63.8\% ($\times$2.01) & 66.7\% ($\times$2.00) & 65.5\% ($\times$1.87) & 67.9\% ($\times$1.78) \\
Police & 58.4\% ($\times$1.84) & 52.9\% ($\times$1.59) & 58.0\% ($\times$1.66) & 63.1\% ($\times$1.66) \\
Migrants & 53.2\% ($\times$1.68) & 59.4\% ($\times$1.78) & 58.7\% ($\times$1.68) & 62.3\% ($\times$1.64) \\
\bottomrule
\end{tabular*}
\caption{Year-by-year any-salience rates for the five highest-salience groups in the full-corpus group audit. Parenthetical values are lifts relative to the corpus-wide any-salience baseline for the same calendar year, which rises from 31.7\% (2022) to 38.1\% (2025). The same three groups, Jews, the Far-right, and Muslims, rank highest in every year, indicating that the main RQ3 pattern is not reducible to a single late-period news cycle.}
\label{tab:yearly_groups}
\end{table*}

\begin{table*}[t]
\centering
\scriptsize
\setlength{\tabcolsep}{6pt}
\renewcommand{\arraystretch}{1.1}
\begin{tabularx}{\textwidth}{@{}lccX@{}}
\toprule
\textbf{Group} & \textbf{UvT lift} & \textbf{UvT rate} & \textbf{Interpretation} \\
\midrule
Far-right & $\times$10.60 & 56.0\% & elevated directional signal only; not a primary prevalence estimate \\
Muslims & $\times$6.83 & 36.1\% & elevated directional signal only; explicit mentions only \\
LFI & $\times$2.97 & 15.7\% & weaker directional signal in party/mobilisation coverage \\
\bottomrule
\end{tabularx}
\caption{Supplementary us-vs-them analysis. Because the us-vs-them head achieved only moderate human--human ($\kappa=.510$) and human--model ($\kappa=.596$) agreement, it is retained as a supplementary indicator rather than a primary prevalence metric. The values above are therefore best read as qualitative profile signals rather than exact point estimates.}
\label{app:uvt_supplement}
\end{table*}

\begin{table*}[t]
\centering
\scriptsize
\setlength{\tabcolsep}{4pt}
\renewcommand{\arraystretch}{1.12}
\begin{tabular*}{\textwidth}{@{\extracolsep{\fill}}lll@{}}
\toprule
\textbf{Hyperparameter} & \textbf{CamemBERT-base} & \textbf{XLM-RoBERTa-large} \\
\midrule
Pretrained model             & \texttt{camembert-base}       & \texttt{xlm-roberta-large} \\
Max sequence length (tokens) & 48                            & 48 \\
Per-device batch size        & 16                            & 8 \\
Gradient accumulation steps  & 2 (eff.\ batch = 32)          & 4 (eff.\ batch = 32) \\
Learning rate                & $2 \times 10^{-5}$            & $1 \times 10^{-5}$ \\
Weight decay                 & 0.01                          & 0.01 \\
Dropout                      & 0.2                           & 0.1 \\
Warmup ratio                 & 0.06                          & 0.06 \\
Max epochs                   & 20                            & 20 \\
Early stopping patience      & 3 (validation loss)           & 3 (validation loss) \\
Optimizer                    & AdamW                         & AdamW \\
Loss (salience/charge heads) & BCE with class-balanced pos.\ weights & BCE with class-balanced pos.\ weights \\
Loss (story-form head)       & Cross-entropy with class weights & Cross-entropy with class weights \\
Classification pooling       & \texttt{[CLS]} token          & \texttt{[CLS]} token \\
Salience architecture        & Shared encoder, 6 independent binary heads & Shared encoder, 6 independent binary heads \\
Story-form architecture      & Separate encoder, 10-class head & Separate encoder, 10-class head \\
Training seed                & 42                            & 42 \\
Hardware                     & Google Colab (A100 / T4)      & Google Colab (A100 / T4) \\
\bottomrule
\end{tabular*}
\caption{Training hyperparameters for CamemBERT-base and XLM-RoBERTa-large on the final \texttt{v6} supervision set. Both models train on the same 6,999-headline training split; thresholds are selected on the 1,500-headline validation split only. Results in Table~\ref{tab:model_comparison} are from the seed-42 run; three-seed evaluation (seeds 42, 123, 456) confirms seed-robust model selection. Full per-seed results: \texttt{v6\_model/colab/multiseed\_variance\_eval.ipynb}.}
\label{app:hyperparams}
\end{table*}

\begin{table*}[t]
\centering
\begin{minipage}[t]{\textwidth}
\centering
\scriptsize
\setlength{\tabcolsep}{3pt}
\renewcommand{\arraystretch}{1.12}
\begin{tabularx}{\textwidth}{@{}lp{0.25\textwidth}p{0.28\textwidth}p{0.26\textwidth}@{}}
\toprule
\textbf{Device} & \textbf{Definition (YES)} & \textbf{Canonical YES examples} & \textbf{Canonical NO examples} \\
\midrule
Loaded vocabulary &
Strong emotional, exaggerated, metaphorical, or value-laden wording beyond neutral reporting &
\textit{catastrophe, chaos, désastre, submersion, effondrement}; metaphorical or dramatic phrasing &
Neutral descriptive wording: \textit{réforme, mesure, discussion} \\
\addlinespace
Blame attribution &
An actor is clearly presented as causing harm, failure, or a negative outcome &
Explicit causation: \textit{provoque, entraîne, aggrave}; responsibility: \textit{responsable de, à cause de}; accusation: \textit{accusé de, mis en cause} &
Simple criticism or opposition without causation: \textit{X critique Y}, \textit{X s'oppose à Y} \\
\addlinespace
Threat framing &
Wording casts an actor, event, or situation as dangerous, destabilizing, or harmful &
\textit{menace, danger, risque, fait craindre}; escalation: \textit{au bord de, sous pression}; metaphorical threat: \textit{déferlante, explosion de, submerge} &
Neutral reporting of events without danger framing \\
\addlinespace
Rhetorical question &
A question that implies judgment, criticism, or suspicion &
Implies evaluation: \textit{Faut-il craindre~? Macron a-t-il échoué~? Jusqu'où ira\ldots} &
Neutral informational questions: \textit{Que dit la loi~? Que sait-on~?} \\
\addlinespace
Us-vs-them$^\dagger$ &
Explicit opposition between two distinct groups, both holding social, political, ethnic, national, class, or institutional identity &
\textit{La France face aux migrants}; class opposition: \textit{les 1\% vs le reste du monde}; institutional: \textit{gouvernement vs syndicats} &
Politician vs.\ politician (CONFLICT); commercial competition; one actor + vague target; crime report victim/perpetrator \\
\addlinespace
Charge band (HIGH) &
HIGH = strong emotional intensity, crisis, violence, confrontation, or scandal &
Crisis, attack, scandal, strong confrontation & Routine, neutral reporting (LOW) \\
\bottomrule
\end{tabularx}
\captionof{table}{Condensed annotation schema for the released supervision criteria. $^\dagger$Us-vs-them requires both groups to hold explicit social, political, ethnic, national, class, or institutional identity; borderline guidance covers politician-vs.-politician (classified as CONFLICT) and commercial competition as canonical exclusions. The full annotation prompt, user template, lexicons, and analysis scripts are available at \url{https://github.com/lefrenchnewslab/framing-wording-selection}.}
\label{app:prompt}
\end{minipage}
\end{table*}

\vspace{0.8em}
\begin{table*}[t]
\begin{minipage}[t]{\textwidth}
\centering
\small
\setlength{\tabcolsep}{10pt}
\begin{tabular*}{0.72\textwidth}{@{\extracolsep{\fill}}lrr@{}}
\toprule
\textbf{Metric} & \textbf{Value} & \textbf{N} \\
\midrule
Cohen's $\kappa$ (nominal)       & .654 & 642 \\
Cohen's $\kappa$ (linear)        & .666 & 642 \\
Krippendorff's $\alpha$          & .653 & 642 \\
\% raw agreement                 & 69.9\% & 642 \\
\midrule
\multicolumn{3}{l}{\textit{By primary annotator confidence}} \\
\quad HIGH                       & .729 & 318 \\
\quad MEDIUM                     & .586 & 314 \\
\quad LOW                        & .157 & 10 \\
\bottomrule
\end{tabular*}
\captionof{table}{Post-hoc human--human inter-annotator agreement on the 642 three-way story-form conflict cases resolved by single-annotator arbitration. A second independent annotator applied the same codebook without access to the primary annotator's decisions. The original gold labels were not modified. Agreement is higher for cases the primary annotator rated HIGH-confidence ($\kappa=.729$) than MEDIUM-confidence ($\kappa=.586$), confirming well-calibrated self-assessed uncertainty. Lower agreement on conflict trios involving ELITE ($\kappa=.276$--.417) reflects the residual-category nature of ELITE in the schema. Interpretation thresholds follow \citet{landisKoch1977}.}
\label{app:arbitration_iaa}
\end{minipage}
\end{table*}

\begin{table*}[tp]
\centering
\small
\setlength{\tabcolsep}{6pt}
\begin{tabularx}{\textwidth}{l l l X}
\toprule
\textbf{Outlet} & \textbf{Default cell} & \textbf{Perturbed cell(s)} & \textbf{Triggering perturbation(s)} \\
\midrule
Valeurs actuelles & double-distinctive & selection-dominant & raise salience median by 10\% or 20\% \\
Le HuffPost & low-distinctiveness & selection-dominant & lower salience median by 10\% or 20\%; any-salience-only perturbation except raising selection median \\
Marianne & selection-dominant & double-distinctive & lower salience median by 10\% or 20\%; any selection-only perturbation \\
Lib\'eration & low-distinctiveness & selection-dominant & lower selection median by 20\% \\
BFMTV & salience-dominant & double-distinctive; low-distinctiveness & lower selection median by 10\% or 20\%; raise salience median by 20\% \\
L'Humanit\'e & selection-dominant & double-distinctive & lower salience median by 10\% or 20\% \\
20 Minutes & salience-dominant & low-distinctiveness & raise salience median by 10\% or 20\% \\
\bottomrule
\end{tabularx}
\caption{Outlets whose four-cell assignments change under $\pm$20\% perturbations of the salience and selection median split points. All remaining outlets keep the same cell assignment under every perturbation, indicating that instability concentrates in near-boundary cases, not distinctive outliers.}
\label{tab:typology_perturbation}
\end{table*}

\begin{table*}[tp]
\centering
\small
\setlength{\tabcolsep}{4pt}
\renewcommand{\arraystretch}{1.08}
\begin{tabularx}{\textwidth}{@{}lrrrrr@{}}
\toprule
\textbf{Outlet} & \textbf{N} & \textbf{Full Any-Sal\%} & \textbf{RQ-excl Any-Sal\%} & \textbf{$\Delta$ pp} & \textbf{$\Delta$ Sal.JS} \\
\midrule
Fdesouche & 16,124 & 72.6 & 72.2 & 0.5 & $-.026$ \\
Slate.fr & 3,907 & 49.6 & 31.9 & 17.7 & $-.032$ \\
Causeur & 3,056 & 59.1 & 50.0 & 9.1 & $+$.002 \\
Blast$^\dagger$ & 646 & 67.3 & 65.9 & 1.4 & $-.018$ \\
Ouest-France & 90,423 & 7.2 & 5.8 & 1.4 & $-.002$ \\
Mediapart & 7,092 & 58.8 & 56.8 & 2.0 & $-.009$ \\
Valeurs actuelles & 14,286 & 52.0 & 49.9 & 2.1 & $-.009$ \\
Les Echos & 55,694 & 24.2 & 21.0 & 3.2 & $+$.004 \\
L'Express & 11,795 & 45.1 & 38.1 & 7.0 & $-.001$ \\
TF1 INFO & 44,014 & 38.3 & 27.4 & 10.9 & $-.019$ \\
20 Minutes & 62,832 & 39.1 & 33.5 & 5.7 & $-.000$ \\
BFMTV & 57,617 & 29.9 & 27.8 & 2.1 & $-.001$ \\
Marianne & 9,412 & 61.2 & 53.1 & 8.0 & $+$.001 \\
JDD & 17,598 & 41.9 & 37.9 & 4.0 & $-.001$ \\
L'Humanit\'e & 17,911 & 49.9 & 44.1 & 5.8 & $+$.001 \\
Le Figaro & 108,307 & 33.2 & 30.3 & 2.9 & $-.001$ \\
Le Parisien & 93,877 & 40.3 & 37.3 & 3.1 & $-.001$ \\
Franceinfo & 83,574 & 31.7 & 27.0 & 4.6 & $-.001$ \\
Le Monde & 39,017 & 37.2 & 34.8 & 2.5 & $-.004$ \\
La Croix & 35,980 & 29.3 & 25.6 & 3.7 & $+$.000 \\
Lib\'eration & 33,921 & 45.7 & 40.9 & 4.8 & $+$.000 \\
Le Point & 32,522 & 40.2 & 33.9 & 6.3 & $-.001$ \\
CNews & 27,252 & 38.5 & 35.0 & 3.5 & $-.001$ \\
Le HuffPost & 17,737 & 37.4 & 32.3 & 5.0 & $-.000$ \\
Le Nouvel Obs & 17,517 & 42.2 & 36.8 & 5.4 & $-.000$ \\
\midrule
\textit{Corpus} & 902,111 & 34.6 & 30.6 & 4.0 & --- \\
\bottomrule
\end{tabularx}
\caption{Rhetorical-question-excluded any-salience robustness. Recomputing any-salience as loaded vocabulary OR blame attribution OR threat framing (dropping the rhetorical-question head) reduces the corpus-level detected rate from 34.6\% to 30.6\% (4.0 pp). The Sal.JS/Sel.JS Pearson correlation drops from $r=.736$ to $r=.561$ ($r=.650$ excluding Fdesouche). Four near-median outlets flip typology cells: Mediapart and Valeurs actuelles (double-distinctive $\rightarrow$ selection-dominant) and Marianne and L'Humanit\'e (selection-dominant $\rightarrow$ double-distinctive); all other 21 outlets are stable. $^\dagger$Blast ($N{=}646$): case profile, wide CIs (Table~\ref{tab:ci_audit}).}
\label{app:rq_excluded}
\end{table*}

\begin{table*}[tp]
\centering
\small
\setlength{\tabcolsep}{4pt}
\renewcommand{\arraystretch}{1.08}
\begin{tabularx}{\textwidth}{@{}lrrrrrr@{}}
\toprule
\textbf{Head} & \textbf{Threshold} & \textbf{Val.\ Prev.} & \textbf{Val.\ Prec.} & \textbf{Prod.\ Prev.} & \textbf{Est.\ Prec.} & \textbf{$\Delta$ pp} \\
\midrule
Loaded vocabulary & 0.67 & 23.9\% & 73.3\% & 17.0\% & 64.3\% & $-$9.0 \\
Blame attribution & 0.83 & 14.7\% & 71.9\% & 13.6\% & 70.1\% & $-$1.8 \\
Threat framing & 0.84 & 8.8\% & 72.3\% & 8.5\% & 71.5\% & $-$0.8 \\
Rhetorical question & 0.89 & 7.3\% & 95.7\% & 4.4\% & 92.9\% & $-$2.8 \\
Us-vs-them & 0.89 & 6.4\% & 60.0\% & 5.3\% & 55.0\% & $-$5.0 \\
\midrule
\textbf{Any-salience} & \textbf{OR} & \textbf{41.1\%} & \textbf{83.8\%} & \textbf{33.7\%} & \textbf{79.0\%} & \textbf{$-$4.8} \\
\bottomrule
\end{tabularx}
\caption{Estimated precision drift under class-prior shift. Expected production precision is computed from the validation-set TPR and FPR at each head's chosen threshold, applying the production corpus prevalence via \citet{elkan2001}. The any-salience OR row is an XLM-R baseline diagnostic computed before the final CamemBERT rhetorical-question substitution, so its production prevalence (33.7\%) is slightly lower than the final ensemble corpus any-salience rate in the main text (34.6\%).}
\label{app:threshold_drift}
\end{table*}

\begin{table*}[t]
\centering
\small
\begin{minipage}[t]{0.49\textwidth}
\centering
\textbf{(a) Sparse-cell exclusion sensitivity}\par\vspace{2pt}
\setlength{\tabcolsep}{4pt}
\begin{tabular}{lrrrr}
\toprule
\textbf{Min.\ headlines} & \textbf{N cells} & \textbf{$\beta$} & \textbf{SE} & \textbf{$p$} \\
\midrule
5  & 1,197 & 0.8793 & 0.0637 & $<$.001 \\
10 & 1,184 & 0.7582 & 0.0660 & $<$.001 \\
20 & 1,161 & 0.8552 & 0.0689 & $<$.001 \\
30 & 1,151 & 0.9218 & 0.0684 & $<$.001 \\
\bottomrule
\end{tabular}
\end{minipage}\hfill
\begin{minipage}[t]{0.49\textwidth}
\centering
\textbf{(b) Cluster-robust inference ($G{=}25$)}\par\vspace{2pt}
\setlength{\tabcolsep}{3.5pt}
\begin{tabular}{lrrrr}
\toprule
\textbf{Specification} & \textbf{$\beta$} & \textbf{Cluster SE} & \textbf{$p$} & \textbf{95\% CI} \\
\midrule
Outlet FE  & 0.862 & 0.443 & .063 & $[-$.05, 1.78$]$ \\
Two-way FE & 0.758 & 0.420 & .084 & $[-$.11, 1.63$]$ \\
\bottomrule
\end{tabular}
\end{minipage}
\caption{Temporal panel robustness. (a) Sensitivity to sparse-cell exclusion thresholds (two-way outlet$+$month fixed effects, classical SEs): the within-outlet coupling is stable across thresholds. (b) Cluster-robust standard errors (clustered by outlet, $G{=}25$, small-sample correction): positive under both specifications but not individually significant, consistent with treating the panel as confirmatory rather than a stand-alone test.}
\label{app:panel_sensitivity}
\label{app:panel_wild}
\end{table*}

\end{document}